\documentclass[10pt,twocolumn,letterpaper]{article}
\usepackage[accsupp]{axessibility}  

\usepackage[pagenumbers]{iccv} 

\usepackage{amsfonts}
\usepackage{amsmath}
\usepackage{graphicx}
\usepackage{longtable}
\usepackage{float}
\usepackage{multirow}
\usepackage{booktabs}
\usepackage{subcaption}
\usepackage{pifont}
\usepackage{xspace}           
\usepackage{xcolor}           

\definecolor{myyellow}{HTML}{DAA520}

\newcommand{\ours}{Mixed Signals\xspace}
\newcommand{\lidar}{LiDAR\xspace}

\newcommand{\cmark}{\ding{51}}%
\newcommand{\xmark}{\ding{55}}%

\newcommand{\evone}{\texttt{EV-1}\xspace}
\newcommand{\evonetable}{EV-1\xspace}
\newcommand{\evtwo}{\texttt{EV-2}\xspace}
\newcommand{\evtwotable}{EV-2\xspace}
\newcommand{\laser}{\texttt{Laser}\xspace}
\newcommand{\lasertable}{Laser\xspace}
\newcommand{\rsutop}{\texttt{TOP}\xspace}
\newcommand{\rsudome}{\texttt{DOME}\xspace}

\newcommand\mypara[1]{\vspace{1.1mm}\noindent\textbf{#1}}

\makeatletter
\let\temp@ESO@isMEMOIR\ESO@isMEMOIR
\let\temp@AtTextUpperLeft\AtTextUpperLeft
\let\temp@AtPageUpperLeft\AtPageUpperLeft
\let\temp@AtPageLowerLeft\AtPageLowerLeft
\let\temp@AtPageCenter\AtPageCenter
\let\temp@AtTextLowerLeft\AtTextLowerLeft
\let\temp@AtTextCenter\AtTextCenter
\let\temp@ESO@HookI\ESO@HookI
\let\temp@ESO@HookII\ESO@HookII
\let\temp@ESO@HookIII\ESO@HookIII
\let\temp@ClearShipoutPicture\ClearShipoutPicture
\let\temp@AddToShipoutPicture\AddToShipoutPicture
\let\temp@ESO@yoffsetI\ESO@yoffsetI
\let\temp@ESO@yoffsetII\ESO@yoffsetII

\let\ESO@isMEMOIR\@undefined
\let\AtTextUpperLeft\@undefined
\let\AtPageUpperLeft\@undefined
\let\AtPageLowerLeft\@undefined
\let\AtPageCenter\@undefined
\let\AtTextLowerLeft\@undefined
\let\AtTextCenter\@undefined
\let\ESO@HookI\@undefined
\let\ESO@HookII\@undefined
\let\ESO@HookIII\@undefined
\let\ClearShipoutPicture\@undefined
\let\AddToShipoutPicture\@undefined
\let\ESO@yoffsetI\@undefined
\let\ESO@yoffsetII\@undefined

\usepackage{pdfpages}

\let\ESO@isMEMOIR\temp@ESO@isMEMOIR
\let\AtTextUpperLeft\temp@AtTextUpperLeft
\let\AtPageUpperLeft\temp@AtPageUpperLeft
\let\AtPageLowerLeft\temp@AtPageLowerLeft
\let\AtPageCenter\temp@AtPageCenter
\let\AtTextLowerLeft\temp@AtTextLowerLeft
\let\AtTextCenter\temp@AtTextCenter
\let\ESO@HookI\temp@ESO@HookI
\let\ESO@HookII\temp@ESO@HookII
\let\ESO@HookIII\temp@ESO@HookIII
\let\ESO@yoffsetI\temp@ESO@yoffsetI
\let\ESO@yoffsetII\temp@ESO@yoffsetII
\makeatother

\definecolor{iccvblue}{rgb}{0.21,0.49,0.74}
\usepackage[pagebackref,breaklinks,colorlinks,allcolors=iccvblue]{hyperref}

\def\paperID{6803} 
\def\confName{ICCV}
\def\confYear{2025}

\title{Mixed Signals: A Diverse Point Cloud Dataset for Heterogeneous LiDAR V2X Collaboration}

\author{
\small{
\textbf{Katie Z Luo}\thanks{Denotes equal contribution.} $^{,1}$\hspace{4pt} 
\textbf{Minh-Quan Dao}\footnotemark[1]   $^{,2,\dagger}$ \hspace{4pt}
\textbf{Zhenzhen Liu}\footnotemark[1] $^{,1}$\hspace{4pt}
\textbf{Mark Campbell}$^{1}$
\hspace{4pt}
\textbf{Wei-Lun Chao}$^{4}$\hspace{4pt}
\textbf{Kilian Q Weinberger}$^{1}$}
\\
\small{
\textbf{Ezio Malis}$^{2}$\hspace{4pt}
\textbf{Vincent Frémont}$^{5}$\hspace{4pt}
\textbf{Bharath Hariharan}$^{1}$\hspace{4pt}
\textbf{Mao Shan}$^{3}$\hspace{4pt}
\textbf{Stewart Worrall}$^{3}$\hspace{4pt}
\textbf{Julie Stephany Berrio Perez}$^{3}$ }
\\
\small{
$^{1}$Cornell University\hspace{4pt}
$^{2}$Inria\hspace{4pt}
$^{3}$University of Sydney\hspace{4pt}
$^{4}$The Ohio State University\hspace{4pt}
$^{5}$École Centrale de Nantes
}
}

\begin{document}
\twocolumn[{%
 \renewcommand\twocolumn[1][]{#1}
 \maketitle

 \vspace{-32px}
\begin{center}
    \captionsetup{type=figure}
    \includegraphics[width=\linewidth]{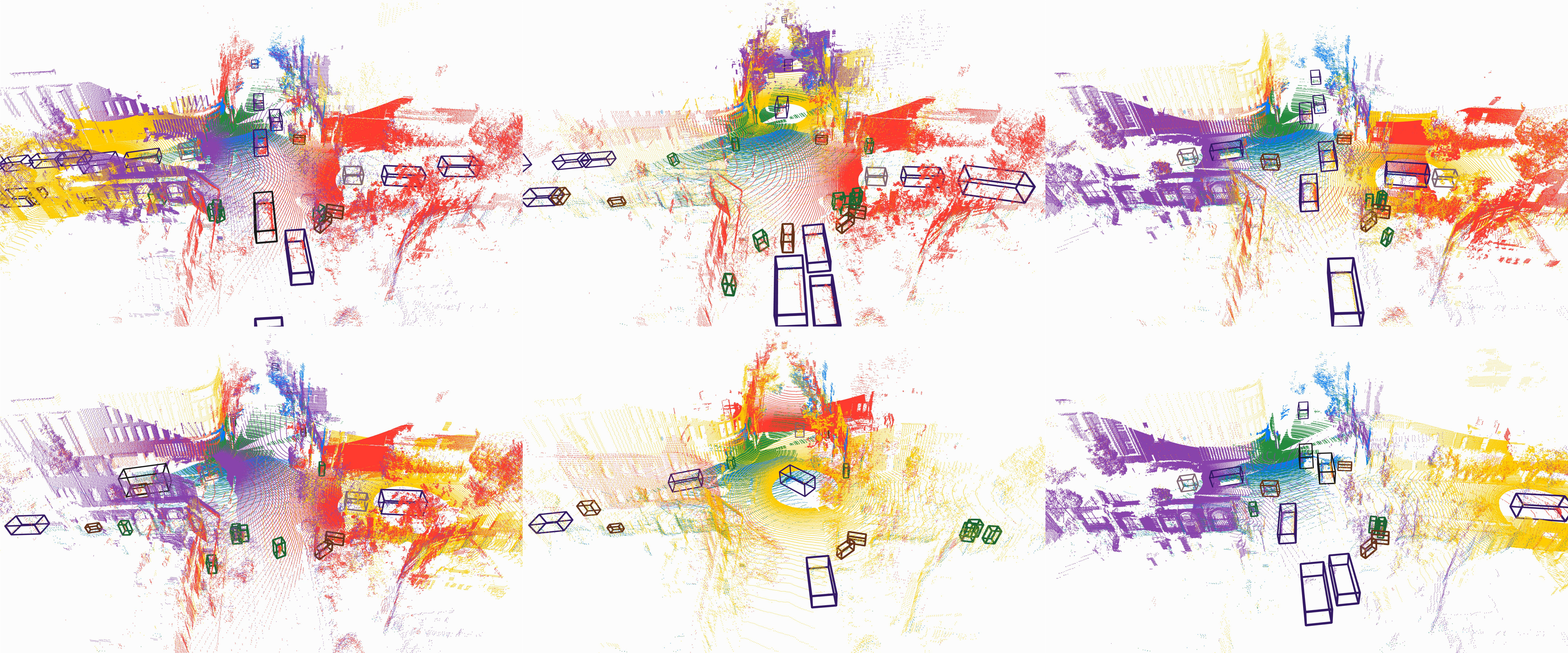}
    \captionof{figure}{\small \textbf{Six samples of the scenes in the dataset.} The different agent's \lidar point clouds are colored as follows: electric vehicle-001 (\evone) in {\color{Purple}purple}, electric vehicle-002 (\evtwo) in {\color{red}red}, urban vehicle (\laser) in {\color{myyellow}yellow}, and the RSU \rsudome and \rsutop{} {\lidar}s in {\color{ForestGreen}green} and {\color{blue}blue}, respectively. We also draw the annotated bounding boxes within the scene. Best viewed in color.}
    \label{fig:teaser}
\end{center}
 }]


 \setcounter{footnote}{0}  

\renewcommand{\thefootnote}{\fnsymbol{footnote}}  
\footnotetext[1]{Denotes equal contribution.}
\footnotetext[2]{Correspondences can be directed to \texttt{minh-quan.dao@inria.fr}.}
\footnotetext[3]{This research is funded by the University of Sydney – Cornell University Ignition Grants/ Global Strategic Collaboration Awards.}

\begin{abstract}

\vspace{-8px}
Vehicle-to-everything (V2X) collaborative perception has emerged as a promising solution to address the limitations of single-vehicle perception systems. However, existing V2X datasets are limited in scope, diversity, and quality.
To address these gaps, we present \ours{}, a comprehensive V2X dataset featuring 45.1k point clouds and 240.6k bounding boxes collected from three connected autonomous vehicles (CAVs) equipped with two different configurations of LiDAR sensors, plus a roadside unit with dual LiDARs. 
Our dataset provides point clouds and bounding box annotations across 10 classes, ensuring reliable data for perception training. We provide detailed statistical analysis on the quality of our dataset and extensively benchmark existing V2X methods on it. \ours{} is \textbf{ready-to-use}, with precise alignment and consistent annotations across time and viewpoints.
We hope our work advances research in the emerging, impactful field of V2X perception. Dataset details at \url{https://mixedsignalsdataset.cs.cornell.edu/}.

\vspace{-10px}
\end{abstract} 

\section{Introduction}
\label{sec:intro}
In recent years, driver assistance~\citep{khan2019comprehensive,nidamanuri2021progressive} and autonomous driving~\citep{yurtsever2020survey,nhtsa_report} technologies have advanced significantly, equipping vehicles with promising capabilities in perception~\citep{lang2019pointpillars,yin2021center}, planning~\citep{hu2022st,hu2023planning}, and control~\citep{amini2020learning,di2021survey}. 
Most of these developments focus on single autonomous vehicle scenarios. Despite the advancements, such settings still face challenges in complex or unpredictable situations~\citep{xu2022opv2v}.
For instance, important traffic participants can be occluded from view, or sensors can fail unexpectedly. As autonomous vehicle deployment increases, new possibilities emerge to address these issues: multiple vehicles can communicate with each other and nearby infrastructure, enabling each vehicle to reliably detect road users even when its own sensors miss them by leveraging shared information. This approach is commonly referred as \textbf{vehicle-to-everything (V2X)} collaborative perception.

\begin{table*}[!ht]
\centering

\resizebox{\linewidth}{!}{%
\begin{tabular}{l c c c c c c c c c c}
\toprule
\multirow{2}{1em}{\textbf{Dataset}}  &
\textbf{Hetero.} &
\textbf{Location} &
\textbf{Driving} &
\textbf{\# Roadside} &
\textbf{\# CAV} &
\textbf{\# Point}  &
\textbf{\# 3D} &
\textbf{\# Classes} &
\textbf{\# Vulnerable} &
\textbf{Track} \\
&
\textbf{Fleet} &
\textbf{} &
\textbf{Side} &
\textbf{LiDARs}&
&
\textbf{Clouds} (K)  &
\textbf{Boxes} (K) &
\textbf{}&
\textbf{Classes} &
\textbf{ID} \\
\midrule
\textcolor{gray}{V2X-Sim \cite{li2022v2x}} & \textcolor{gray}{\xmark} & \multirow{3}{*}{\textcolor{gray}{\shortstack{CARLA\\(Sim.)}}} & \textcolor{gray}{Right} & \textcolor{gray}{1} & \textcolor{gray}{5} & \textcolor{gray}{10.0} & \textcolor{gray}{26.6} & \textcolor{gray}{1} & \textcolor{gray}{0} & \textcolor{gray}{\cmark} \\
\textcolor{gray}{OPV2V \cite{xu2022opv2v}} & \textcolor{gray}{\xmark} & & \textcolor{gray}{Right} & \textcolor{gray}{0} & \textcolor{gray}{2-7} & \textcolor{gray}{11.4} & \textcolor{gray}{232.9} & \textcolor{gray}{1} & \textcolor{gray}{0} & \textcolor{gray}{\xmark} \\
\textcolor{gray}{V2X-Set \cite{xu2022v2x}} & \textcolor{gray}{\xmark} & & \textcolor{gray}{Right} & \textcolor{gray}{2-7} & \textcolor{gray}{2-7} & \textcolor{gray}{33.0} & \textcolor{gray}{230.0} & \textcolor{gray}{1} & \textcolor{gray}{0} & \textcolor{gray}{\xmark} \\
\midrule
DAIR-V2X-C$^{\ddagger}$ \cite{yu2022dair} & \xmark & China & Right & 2 & 1 & 39.0 & 464.0 & 10 & 4 & \xmark \\
V2X-Seq (SPD)$^{\ddagger}$ \cite{yu2023v2x} & \xmark & China & Right & 2 & 1 & 15.0 & 10.4 & 10 & 4 & \cmark \\
RCooper$^{\ddagger}$ \cite{hao2024rcooper}  & \xmark & China & Right & 3 & 0 & 30.0 & \textit{N/A} & 10 & 3 & \xmark \\
HoloVIC$^{\ddagger}$ \cite{Ma_2024} & \xmark & China & Right & 2 & 1 & 100.0 & 1800 & 3 & 2 & \cmark \\
\midrule
Open Mars \cite{Li_2024_CVPR} & \xmark & USA & Right & 0 & 2-3 & 15.0 & 0 & \textit{N/A} & \textit{N/A} & \xmark \\
V2V4Real \cite{xu2023v2v4real} & Height & USA & Right & 0 & 2 & 20.0 & 240.0 & 5 & 0 & \cmark \\
V2X-Real \cite{xiang2024v2x} & Height & USA & Right & 2 & 2 & 33.0 & 1200.0 & 10 & 2 & \cmark \\
\midrule
TUMTrafV2X \cite{zimmer2024tumtraf} & \xmark & Germany & Right & 2 & 1 & 2.0 & 30.0 & 8 & 3 & \cmark \\
\midrule
\textbf{Mixed Signals} & Height, Tilt & AUS & Left & 2 & 3 & 45.1 & 240.6 & 10 & 4 & \cmark \\
\bottomrule
\end{tabular}}
\caption{\small \textbf{Comparison of \ours and existing V2X datasets.} To our best knowledge, \ours is the first dataset to include heterogeneous CAV \lidar configurations, and also the first one that is collected in a left-hand driving country. It captures complex, real-world traffic scenarios and features a diverse range of traffic participants. Those marked with $\ddagger$ are valuable datasets, but are only accessible from certain geographical regions.
}
\label{tab:dataset-compare}
\end{table*}

While single-vehicle perception datasets are abundant across diverse driving conditions~\citep{kitti,lyft,waymo,nuscenes,argoverse,once,ithaca365,pandaset,apolloscape,oxfordrobotcar,cadc}, real-world V2X datasets remain limited in availability, diversity, and quality. Only a handful of publicly available V2X datasets exist~\citep{xu2023v2v4real,xiang2024v2x,zimmer2024tumtraf,Li_2024_CVPR}, with some of them accessible only within specific geographical regions~\citep{yu2022dair,yu2023v2x,hao2024rcooper,Ma_2024}. These data are collected exclusively from three right-hand traffic locations, overlooking the unique traffic dynamics in left-hand traffic countries which make up about a third of the world~\citep{xu2023left}. 
Furthermore, as collaborative perception becomes more widespread, it is valuable for vehicles equipped with different sensor configurations to communicate.
However, in prior datasets, the connected autonomous vehicles (CAVs) share identical or very similar \lidar configurations. 
Finally, as the V2X setting involves multiple agents and sensors, data collection and alignment present additional challenges. 
Often times, difficulty with pose estimation and faulty localization systems result in poor alignment (\autoref{fig:localization-quality}). 
Such inaccuracies can lead to suboptimal performance for detector training~\citep{xu2022v2x}.

To address these limitations, we introduce the \ours dataset, designed to support diverse real-world V2X research scenarios with clean, high-quality data. Notably, \ours is the first V2X dataset that provides heterogeneous CAV \lidar configurations in both position and orientation, and features a left-handing traffic country, Australia. The dataset includes 45.1k point clouds and 240.6k bounding boxes, collected from three CAVs equipped with two configurations of \lidar sensors, along with a roadside unit with two {\lidar}s. It captures a diverse range of traffic participants across 10 different classes, including 4 vulnerable road user categories. Furthermore, compared to existing datasets, \ours offers significantly more precise alignment and consistent annotations across time and viewpoints. 
We emphasize that our dataset is \textit{\textbf{ready-to-use}}; a subset is provided in the supplementary materials, along with the corresponding video visualization showcasing the quality of our collected data and annotations. 
To summarize, our contributions are:
\begin{itemize} 
    \item We introduce the \ours dataset, a high quality, large-scale, publicly available V2X dataset created through careful processing and precise annotations.
    \item To the best of our knowledge, we are the first real-world V2X dataset that encompasses CAV \lidar configurations that differ in both position and orientation, as well as left-hand traffic scenarios.
    \item We provide detailed analysis of the dataset's statistics, and conduct comprehensive benchmarking of existing V2X methods across various settings.
\end{itemize}

\section{Related Works}

While existing collaborative perception datasets have the same sensor setup for their CAVs, our dataset contains three vehicles with two different sensor configurations, including the height and tilt of LiDAR and the type of vehicle.
This difference introduces heterogeneity to our fleet of vehicles, thus making our data more closely resemble the real-world collaboration deployment.
To the best of our knowledge, we have the largest fleet of CAVs with the most diverse sensors of any prior works.

\begin{figure*}
\centering
\begin{subfigure}[]{0.3\linewidth}
\centering
\includegraphics[trim={0cm 0.2cm 0cm 0cm},clip,width=\columnwidth]{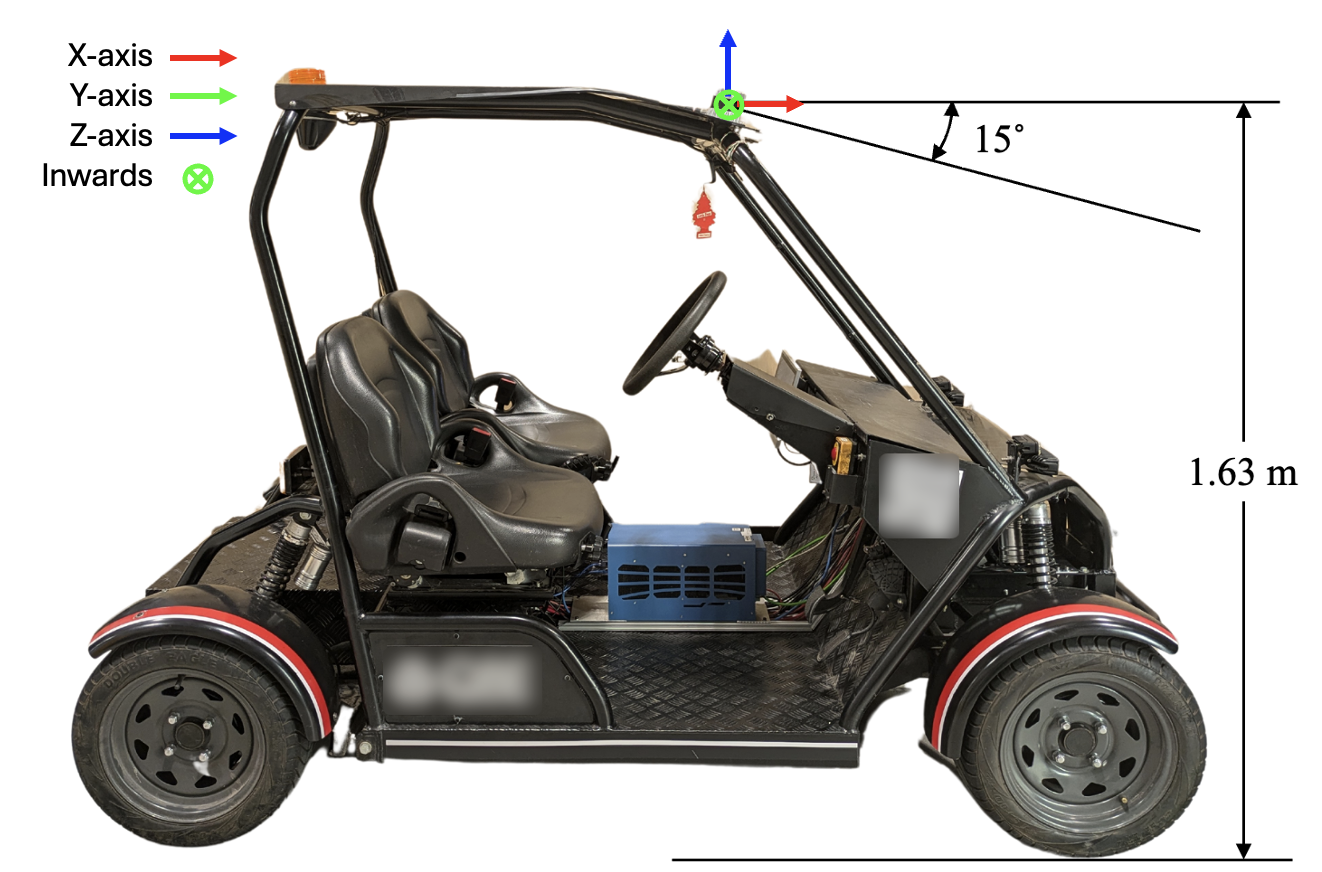}
\label{fig:ev}

\vspace{-8px}
\caption{\small Electric vehicle (EV).}
\label{fig:evs}
\end{subfigure}
\quad
\begin{subfigure}[]{0.4\linewidth}
\centering
\includegraphics[trim={0cm 0cm 0cm 0cm},clip,width=\columnwidth]{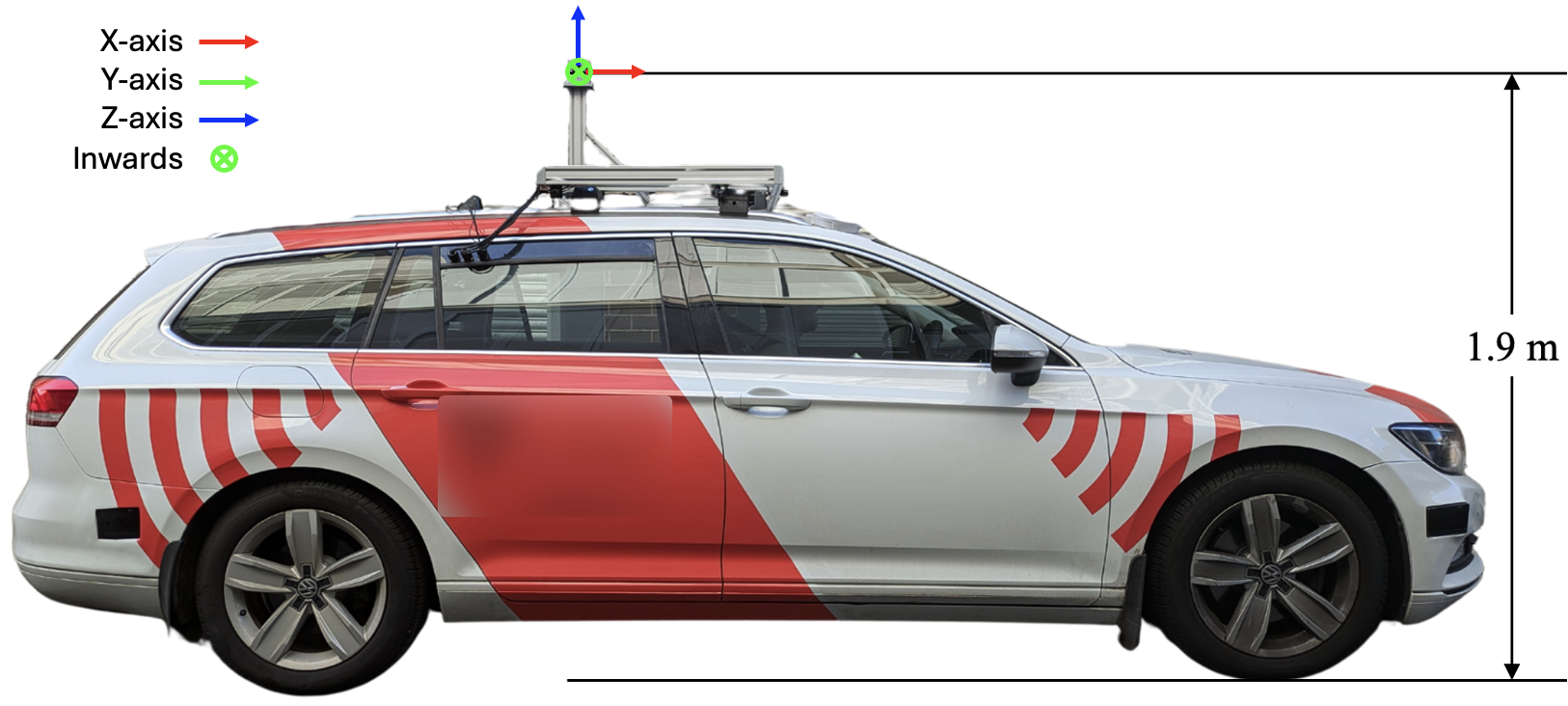}
\label{fig:passat}
\caption{\small Urban vehicle with a OS1-128 beams \lidar.}
\end{subfigure}
\quad
\begin{subfigure}[]{0.2\linewidth}
\centering
\includegraphics[width=\columnwidth]{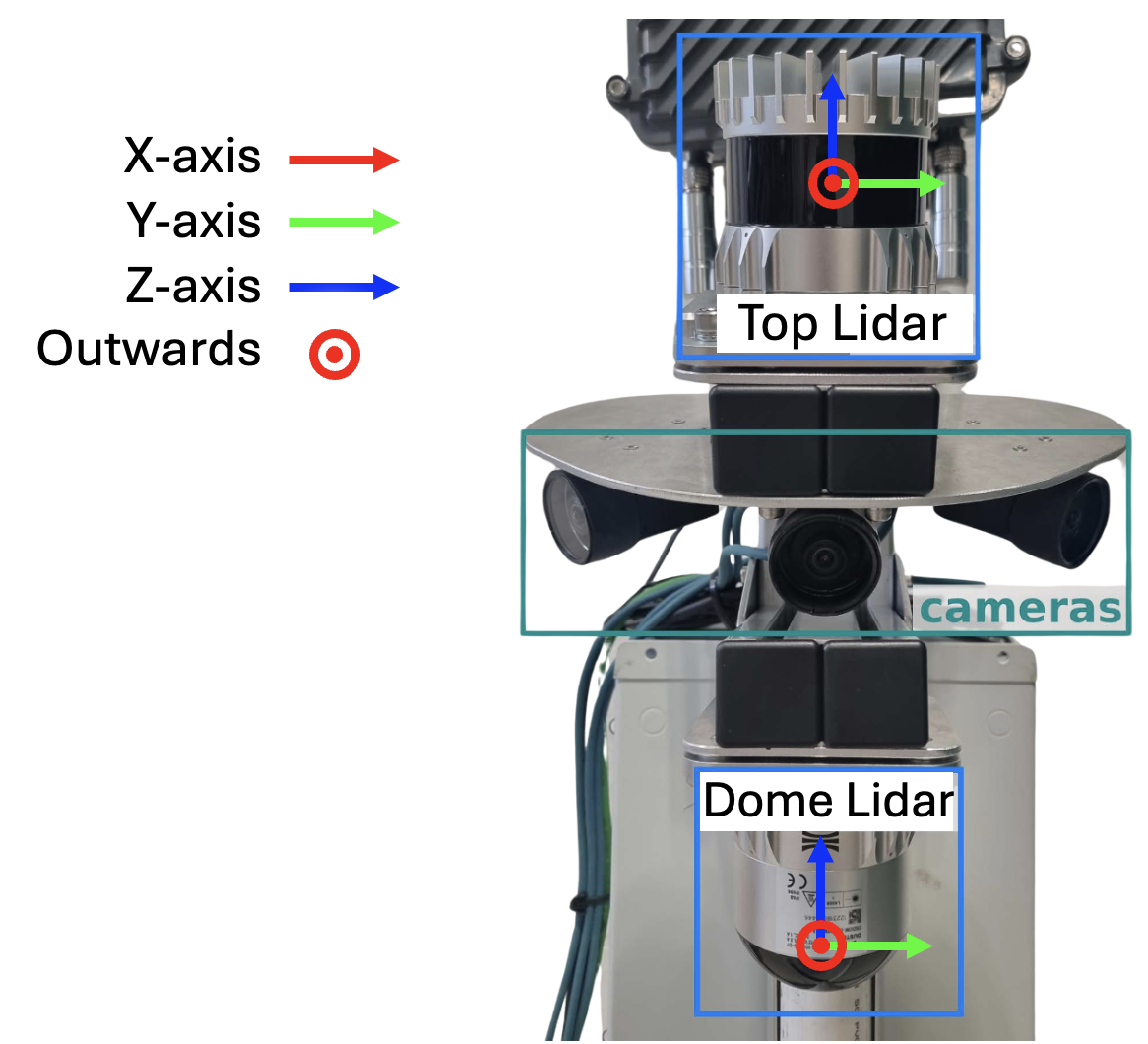}
\label{fig:rsu}
\caption{\small Roadside Unit (RSU).}
\end{subfigure}

\vspace{-8px}
\caption{\small \textbf{Vehicles used for data collection.} (a) is a small electric vehicle outfitted with an OS1-128 beams \lidar system. The \lidar is mounted at a 15° angle relative to the vehicle's body and stands at a height of 1.63 meters. (b) is an urban vehicle equipped with an OS1-128 beam \lidar system located at a height of 1.9 meters. (c) is the RSU which consists of two LiDARs: an OS1-64 beam (\rsutop) and an OSDome-128 (\rsudome) LiDAR mounted on a pole at the intersection at a height of 2.5 meters.}
\label{fig:vehicles}

\vspace{-10px}
\end{figure*}

\mypara{Vehicle-to-Everything Communication.}
One of V2X's objectives is to enhance the perception capabilities of CAVs, facilitating their deployment in urban environments. 
These areas usually have a high presence of Vulnerable Road Users (VRUs), which are defined as people not inside vehicles \cite{def_vru}. 
Despite this, VRUs are under represented in prior works.
The three synthetic datasets made with CARLA \cite{dosovitskiy2017carla} and the real-world dataset V2V4Real \cite{xu2023v2v4real} do not have VRUs.
DAIR-V2X-C \cite{yu2022dair} and its extension V2X-Seq (SPD) \cite{yu2023v2x} provide annotations for 4 VRU classes (pedestrian, bicyclist, tricyclist, and motorcyclist).
However, the absence of details on class distribution in their publications make it hard to judge their VRU coverage.
Additionally, restricted access to these datasets outside China limits their usability.
TUMTrafV2X \cite{zimmer2024tumtraf} annotates 3 VRU classes including pedestrian, bicycle, and motorcycle, which together account for only 24.6\% of the total annotations.
Such underrepresentation causes VRU detection to be overlooked in several collaborative perception studies \cite{wang2020v2vnet, li2021learning, xu2022opv2v, xu2022v2x}.

\mypara{Real World Vehicle-to-Everything Datasets.}
The recent V2X-Real \cite{xiang2024v2x} has a large number pedestrian annotations, which is higher than annotations of the class car, and 3 other VRU classes (scooter, motorcycle, and bicycle).
A drawback of this dataset for VRU detection evaluation is that its benchmark only accounts for pedestrians.
Our dataset contains the highest number of VRU classes, including pedestrian, bicycle, portable personal mobility, and motorcycle.
More importantly, these classes account for 50.3\% of our dataset's total bounding boxes.
Instead of selecting certain VRU classes for benchmarking, we group 4 VRU classes into 2 detection classes as in 
\autoref{sec:dataset_analysis}
to provide a better understanding of how different collaboration methods perform in detecting VRUs. We provide a detailed comparison of our dataset, \ours, with prior works in \autoref{tab:dataset-compare}.

\section{Mixed Signals Dataset}

In this section, we describe the data collection process of the \ours dataset.
We provide a devkit and our full dataset for download on our website: \url{https://mixedsignalsdataset.cs.cornell.edu/}. 

\vspace{-2px}
\subsection{Hardware}

The data collection was carried out using three vehicles and a roadside unit. 

\mypara{Vehicles.}
The three vehicles included two small electric vehicles (EVs) and one urban vehicle, each equipped with OS1 128-beam {\lidar}s, as shown in \autoref{fig:vehicles}. 
The \lidar on the urban vehicle is located horizontally with respect to the ground, while for the EV, the \lidar is tilted downwards 15 degrees. We transformed both EVs' point clouds to have a horizontal reference frame as shown in \autoref{fig:evs}. Although all the vehicles are equipped with the same type of LiDAR sensor, their configurations differ in terms of sensor position and orientation. This variation introduces additional complexity, creating a domain gap between the data collected from different vehicles.

\begin{figure}[t]
\centering
\includegraphics[width=0.99\columnwidth]{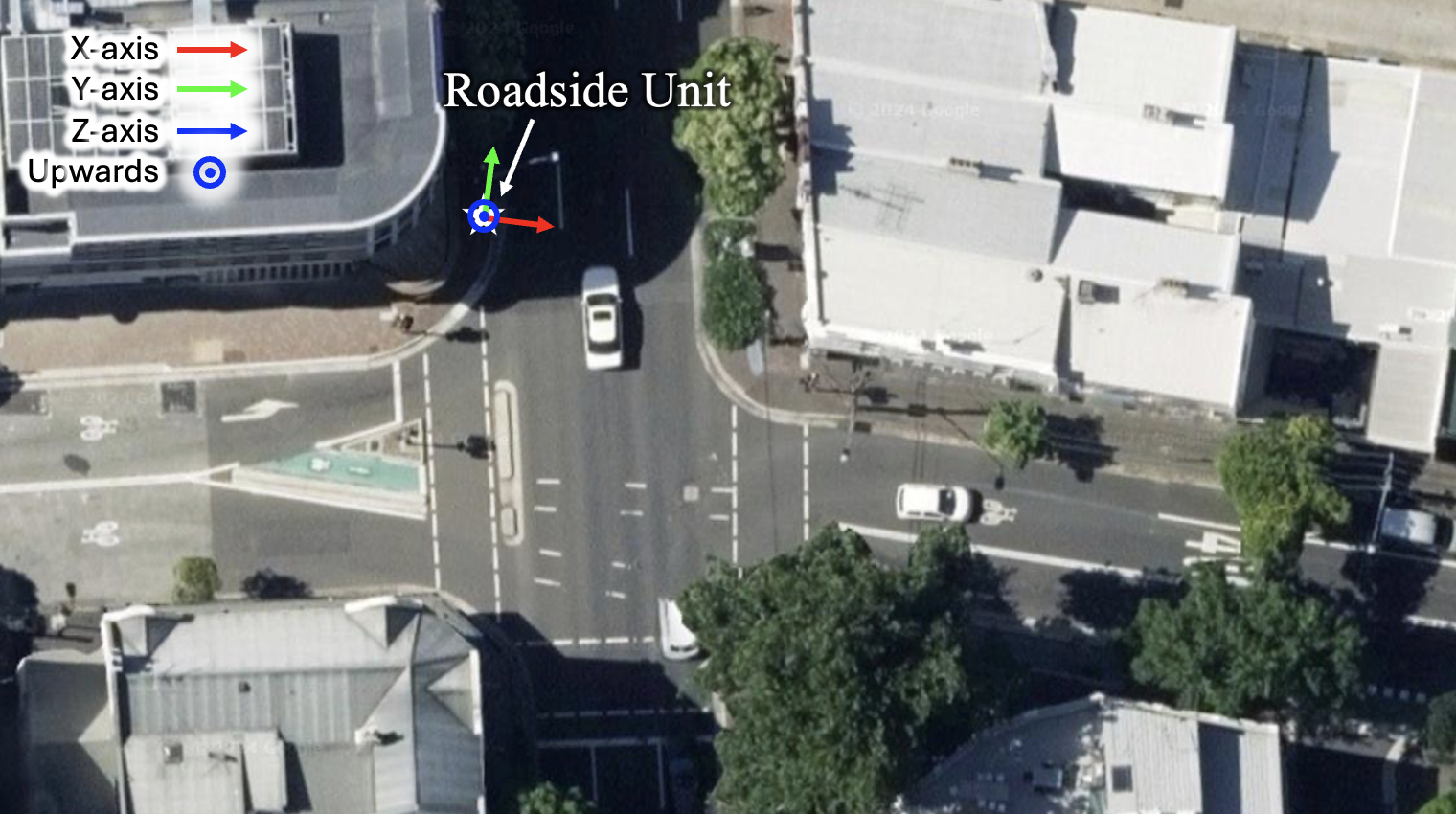}
\caption{\small \textbf{Geographical location of the roadside unit.}}
\label{fig:rsu_location}
\vspace{-16px}
\end{figure}

\mypara{Roadside Unit.}
The roadside unit is equipped with two different LiDAR sensors: an OS-Dome 128-beam for long-range detection and an OS1 64-beam LiDAR for detecting nearby objects. It was located at a fixed geographical position, 2.5 meters above the ground.
The intersection where the roadside unit was installed experiences moderate vehicular traffic and features pedestrian crosswalks along with a bike lane that crosses the intersection. This setup allows us to capture diverse agents during data collection. The placement of the roadside unit is illustrated in \autoref{fig:rsu_location}.

\subsection{Data Acquisition}

The data collection took place at the 
intersection between Abercrombie Street and Myrtle Street in Sydney, Austria,
where the roadside unit is located. The vehicles recorded LiDAR data for two hours during peak rush hour. Throughout this period, the three vehicles repeatedly passed through the intersection. This allowed them to capture interactions between the vehicles and other agents on the road, such as pedestrians, cyclists, and other vehicles. 

\subsubsection{Synchronization and Localization}

\begin{figure}
    \centering
    \includegraphics[width=\columnwidth]{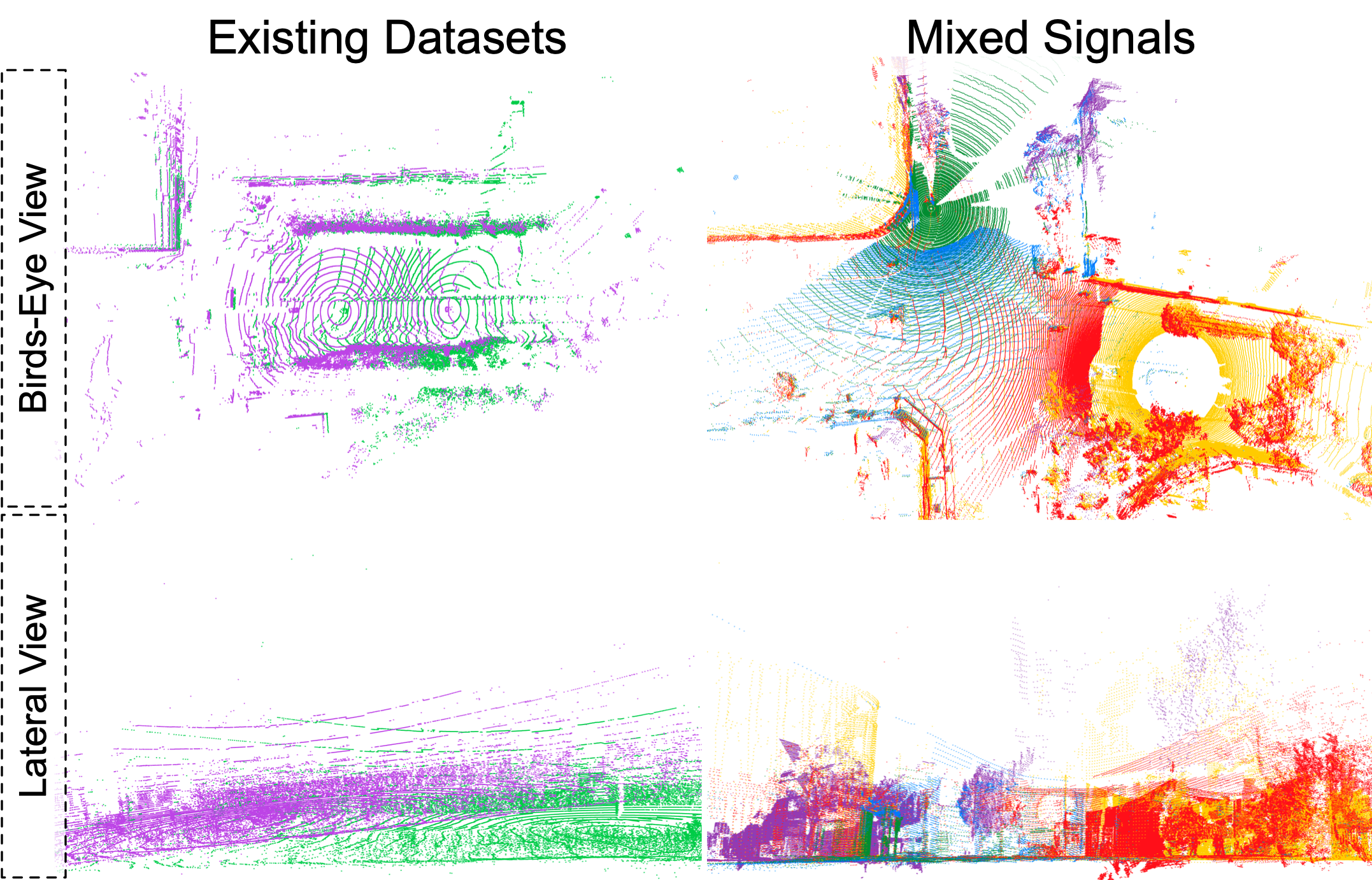}

    \vspace{-4px}
    \caption{\small \textbf{Localization and synchronization quality of \ours and existing datasets.} Different colors correspond to different sensors. In the lateral view, existing datasets visually exhibit vertical inconsistencies, where one point cloud is tilted due to localization errors. In contrast, point clouds in \ours are all accurately aligned.
    }
    \label{fig:localization-quality}

    \vspace{-8px}
\end{figure}

Synchronization and localization are crucial for cross-sensor point cloud alignment. Our dataset employs proven techniques from robotics to achieve precise sensor synchronization and agent localization.
The end result is superior point cloud alignment compared to previous V2X datasets (\autoref{fig:localization-quality}). We describe the details below.

\mypara{Synchronization} refers to the temporal alignment of data streams, ensuring that synchronized sensors capture the same events simultaneously within their overlapping fields of view (FOV). 
We use GPS time to timestamp point clouds captured by our LiDARs at a frequency of 10 Hz.
Even if two vehicles are GPS-synchronized, cross-sensor synchronization still needs to be considered. For example, since the \lidar{} scans the environment in a rotating fashion, the data collected at different spatial locations are captured at slightly different moments.
We defined data samples by setting a time window to match the closest available timestamps from each LiDARs. A maximum timestamp mismatch of 50 milliseconds between point clouds was set to achieve minimal spatial discrepancies. For additional details, refer to Appendix~\ref{sec:apdx-hw-synchronization}.

\mypara{Localization}, i.e., estimating vehicle position relative to a global reference frame, is one of the most critical tasks for CAV.
To overcome inherent problems of Global Navigation Satellite System (GNSS) in urban environments, we use dense and accurate point cloud maps \cite{liosam2020shan} as references for our localization algorithm. 
Both the vehicles and the roadside units are localized within a common reference frame, referred to as the $map\_frame$, which serves as the origin of our map.
The localization algorithm employs a scan-matching technique \cite{ndt} to estimate the vehicles' poses within this map, achieving a maximum positioning error of 15 cm and a heading error of 0.4 degrees.
This allows for consistent spatial alignment between the vehicles and the roadside infrastructure. The vehicles' localization estimates their positions within the $map\_frame$, while the roadside unit is static.
We leave details about map construction and usage in Appendix~\ref{sec:apdx-localization}.

\subsubsection{Scene Selection}

In total, 37 scenes --each consisting of a 30-second snippet-- were carefully selected for inclusion in the dataset due to their rich diversity of vehicles, pedestrians, and cyclists.
The primary goal was to capture various vehicles and vulnerable road users. These scenes encompass a broad spectrum of interactions, including between different types of vehicles and between vehicles and vulnerable road users.
The selected scenes feature intersections of the FOV of the LiDARs of 3 vehicles and the RSU.
Among 37 scenes of our dataset, we select 33 scenes for training and 4 scenes of testing.
The size of the training set and test set are 9553 and 1164 data samples, respectively.
Our selection ensures that there is no temporal overlap between the training set and test set and among scenes of the test set.

\subsection{Dataset Annotation}

The task of 3D object detection for autonomous vehicles requires annotations in the form of 3D bounding boxes, usually parameterized by the center location, three dimensions (length, width, height), and rotation (represented as a quaternion).
To generate such annotations for each data sample, we first aggregate the point clouds of every agent in the coordinate of the roadside unit's top (\rsutop) \lidar{} to focus the annotators' attention to the intersection of interest.
Then, professional annotators from FlipSideAI \cite{flipside} employ the SegmentsAI \cite{segments} annotation tool to label objects and localize them with a 3D bounding box. Classes labeled belong to 10 categories, consisting of: car, truck, pedestrians, bus, electric vehicle, trailer, motorcycle/bike, bicycle, portable personal mobility, and emergency vehicle. 
\autoref{fig:teaser} depicts the annotations applied to the dataset, where each object is enclosed within a cuboid. 

\begin{figure}
    \centering
    \includegraphics[width=\columnwidth]{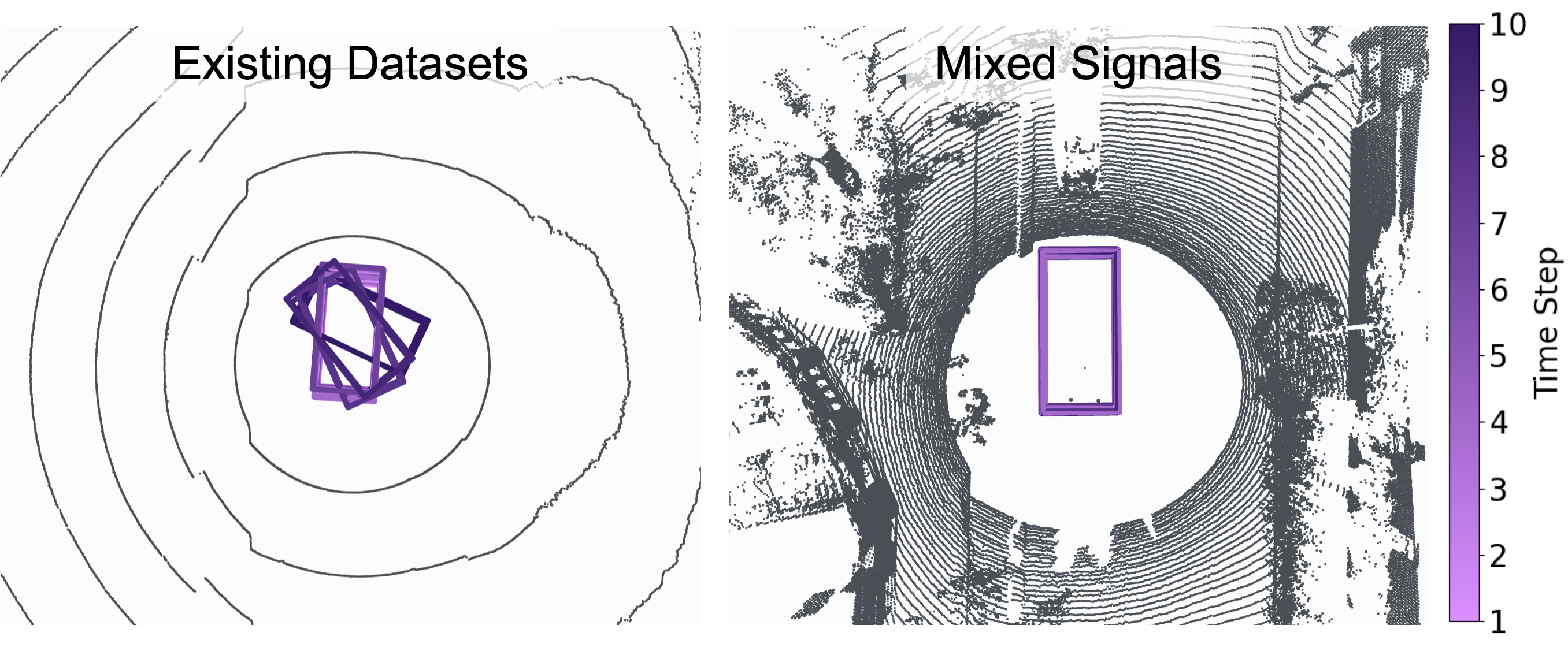}
    \caption{\small \textbf{Label quality of \ours and existing datasets.} We aggregate labels of an object across a entire snippet. Labels in \ours are consistent across time steps and viewpoints.}

    \vspace{-8px}
    \label{fig:label-quality}
\end{figure}

\mypara{Annotations.} Our annotation process involved cycles of monitoring, reviewing, and adjusting labels to meet defined quality objectives.
This allows \ours dataset to extend the quality of the pioneering datasets in the field, which are generally labeled by lay annotators, as shown in \autoref{fig:label-quality}. Here, we reproject the bounding box of a vehicle, as observed from other sensors, back onto its coordinate frame to visualize label consistency. 
Details of the class descriptions and labeling instructions are presented in Appendix Sec.~
\ref{sec:apdx-annotation-instructions}.
While agents in our dataset are synchronized at 10 Hz, we sample keyframes at 1 Hz for manual annotation.
To obtain annotations in a non-key frame, we linearly interpolate the pose of annotations of its closest preceding and succeeding keyframes based on their timestamp.

\vspace{-8px}

\paragraph{Category Labels.}
The Mixed Signals dataset categories consist of road agents in 10 categories of vehicle types and pedestrians including: Car, Truck, Emergency Vehicle, Bus, Motorcycle, Motorized Bike, Portable Personal Mobility Vehicle, (traditional) Bicycle, Electric Vehicle, Trailer, and Pedestrian.
Detailed definition of each category can be found in the appendix.

\subsection{Dataset Analysis}  \label{sec:dataset_analysis}

\begin{figure}[t!]
\centering
\includegraphics[width=\columnwidth]{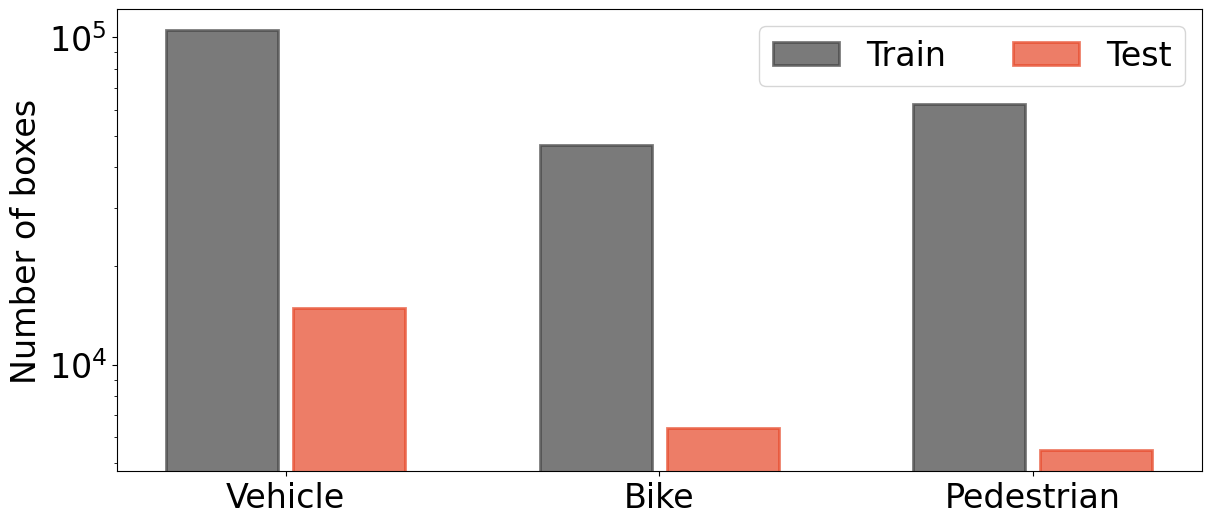}

\vspace{-6px}
\caption{\small \textbf{Number of objects by class.} The y-axis is in log scale.}
\label{fig:num_obj_per_class_in_train_test}

\vspace{-8px}

\end{figure}

\begin{figure*}[t!]
    \centering
    \begin{subfigure}{.32\textwidth}
        \centering
        \includegraphics[width=0.95\linewidth]{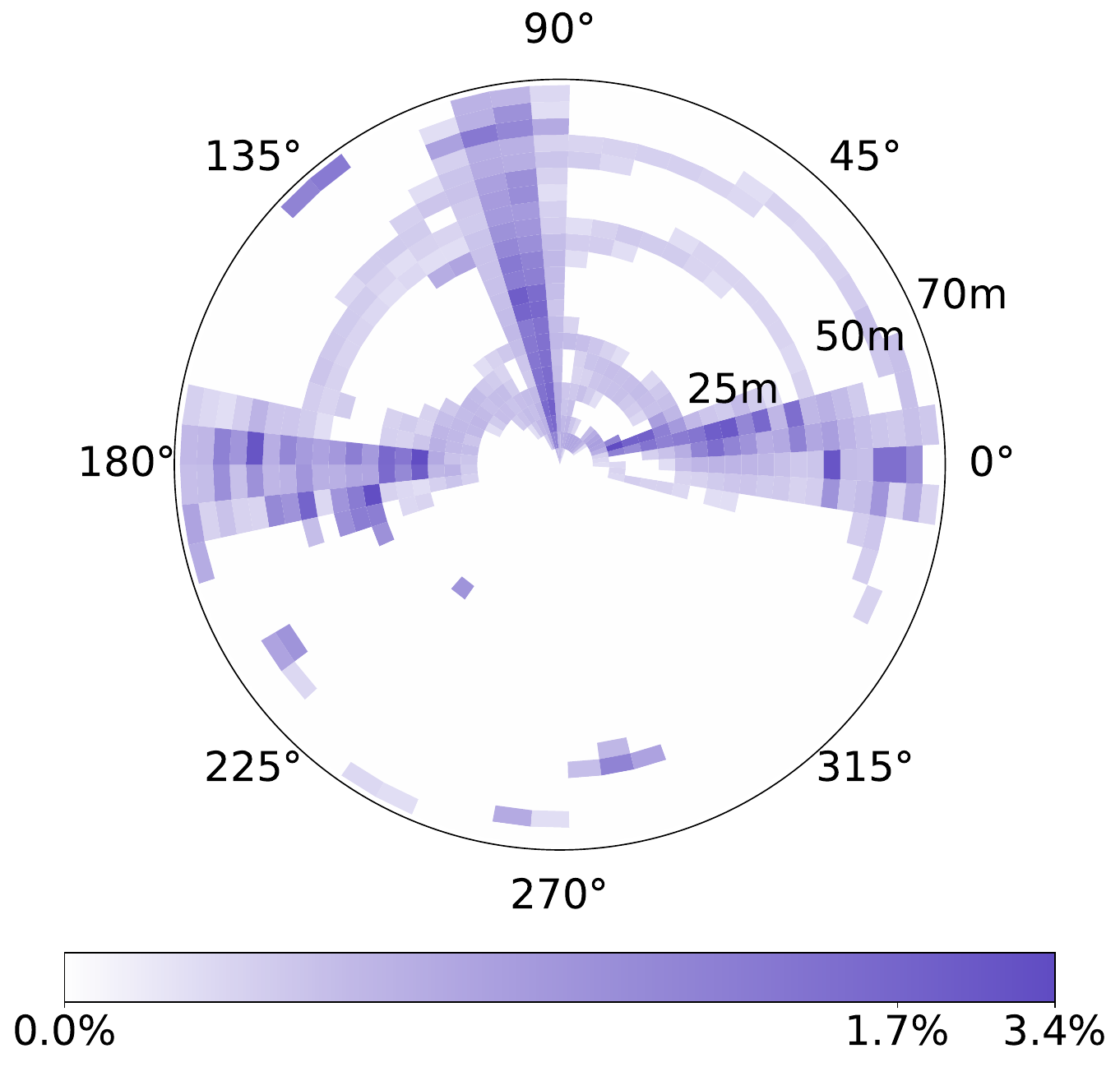}
        \caption{Vehicle}
    \end{subfigure}
    \begin{subfigure}{.32\textwidth}
        \centering
        \includegraphics[width=0.95\linewidth]{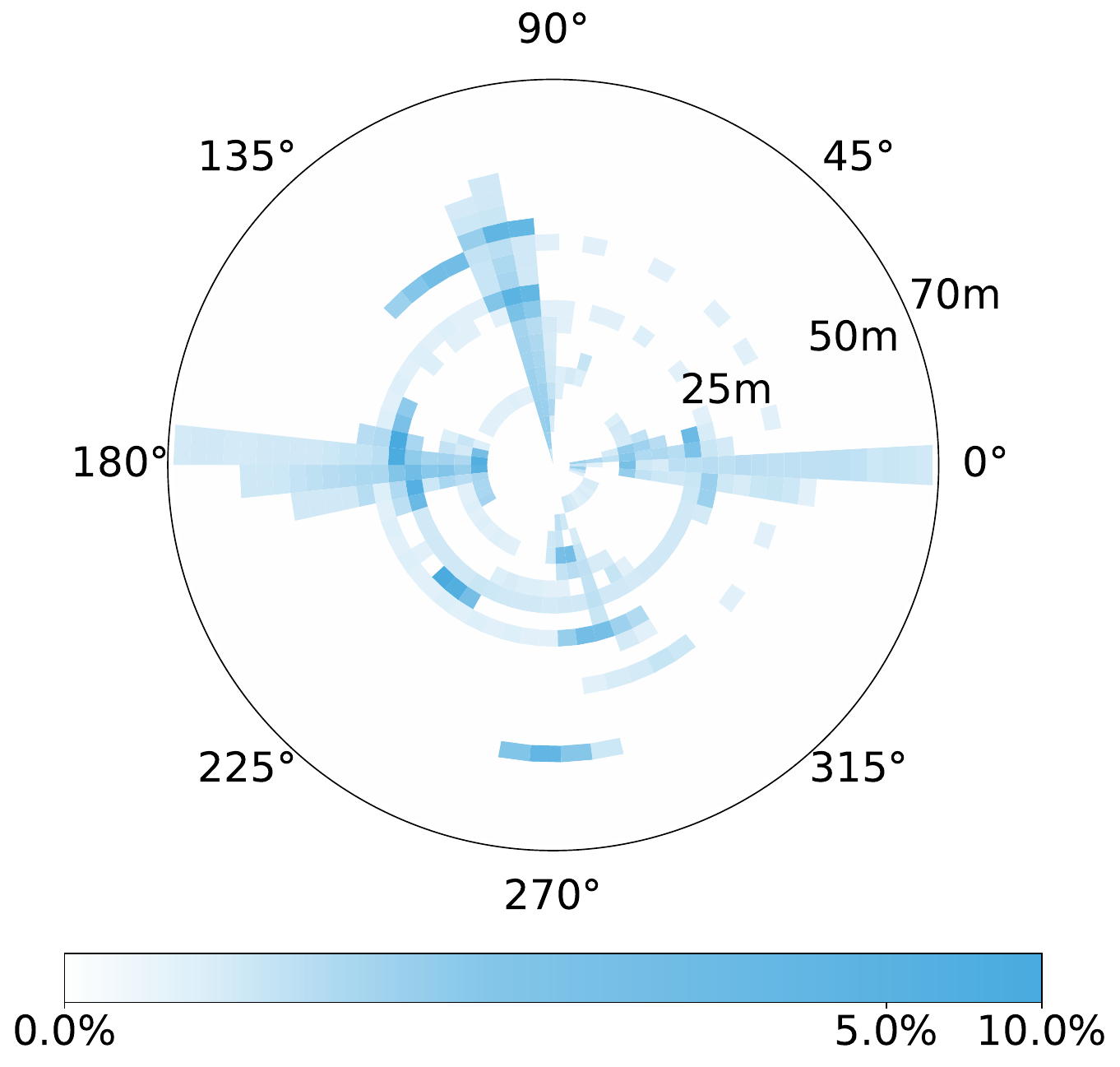}
        \caption{Bike}
    \end{subfigure}
    \begin{subfigure}{.32\textwidth}
        \centering
        \includegraphics[width=0.95\linewidth]{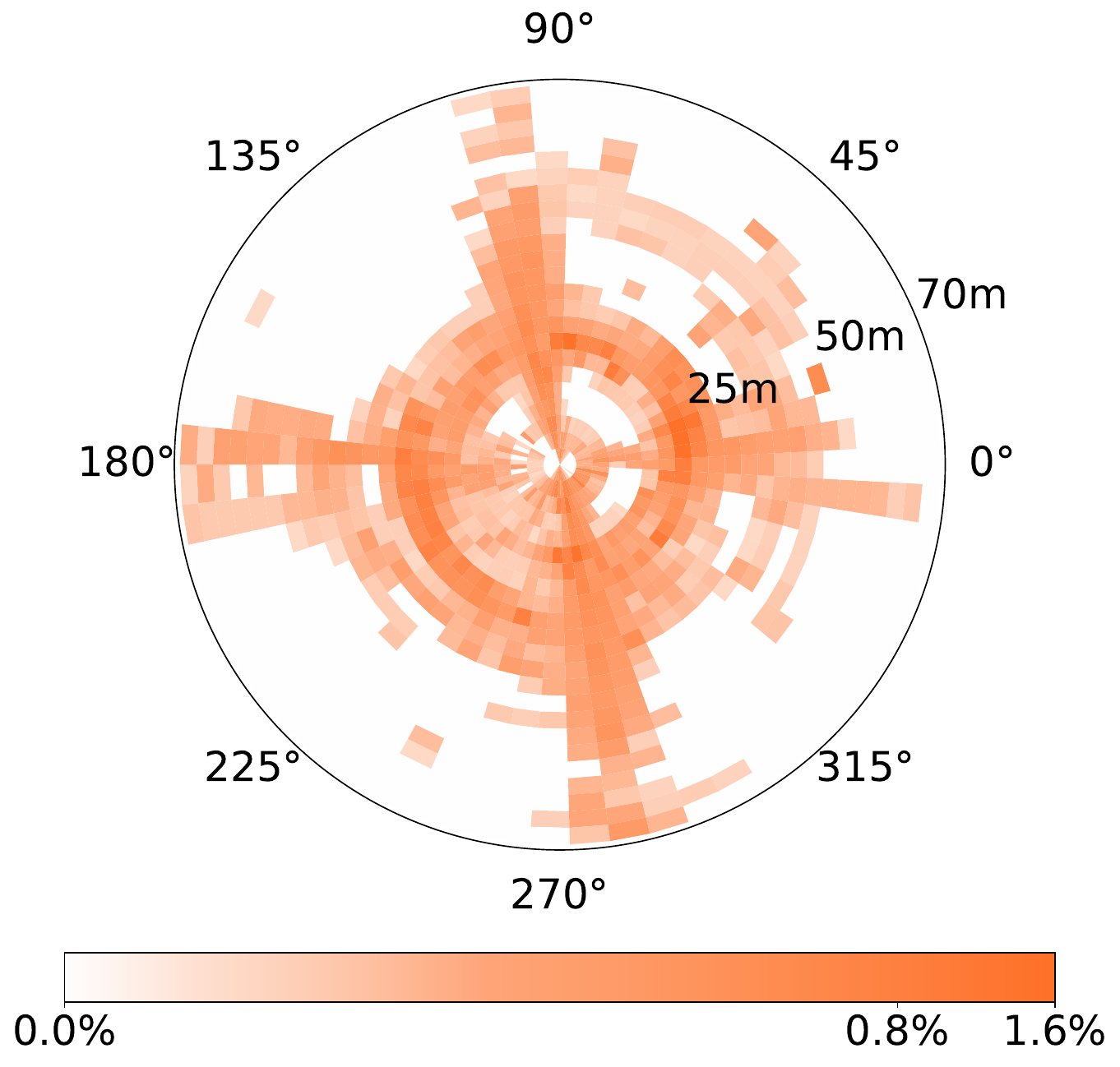}
        \caption{Pedestrian}
    \end{subfigure}%

    \vspace{-8px}
    \caption{\small \textbf{Distribution of annotated object locations.} Locations are shown in polar coordinates relative to the RSU \rsutop sensor.}
    \label{fig:percentage_each_class_in_polar_coord}
\end{figure*}

\begin{figure*}[t!]
    \centering
    \begin{subfigure}{.24\linewidth}
        \centering
        \includegraphics[width=0.95\linewidth]{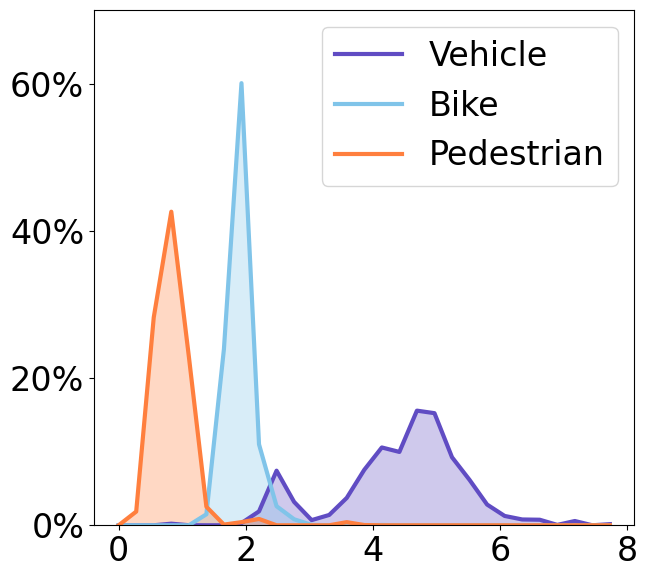}
        \caption{Length (m)}
    \end{subfigure}
    \begin{subfigure}{.24\linewidth}
        \centering
        \includegraphics[width=0.95\linewidth]{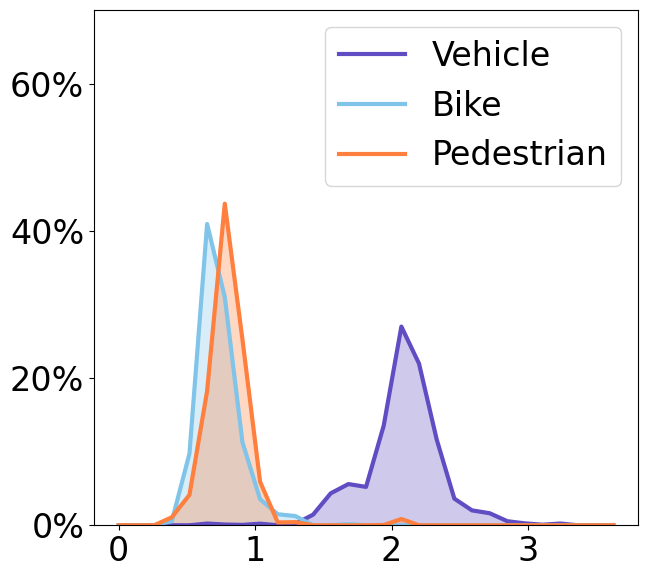}
        \caption{Width (m)}
    \end{subfigure}
    \begin{subfigure}{.24\linewidth}
        \centering
        \includegraphics[width=0.95\linewidth]{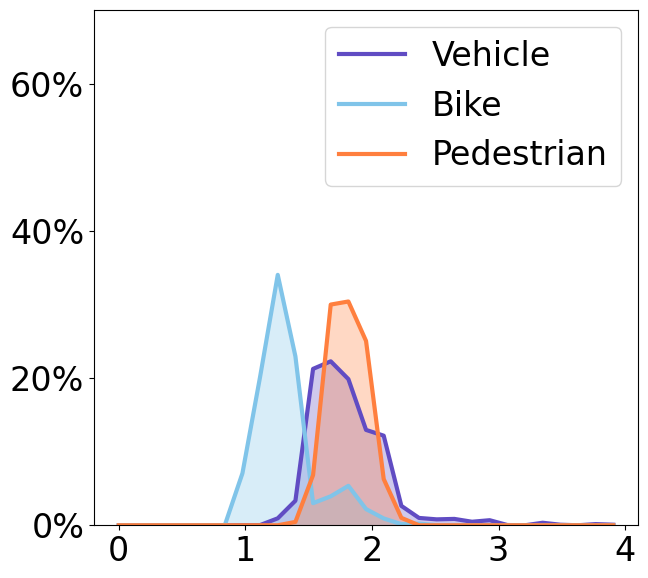}
        \caption{Height (m)}
    \end{subfigure}%
    \begin{subfigure}{.24\linewidth}
    \centering
    \includegraphics[width=0.95\linewidth]{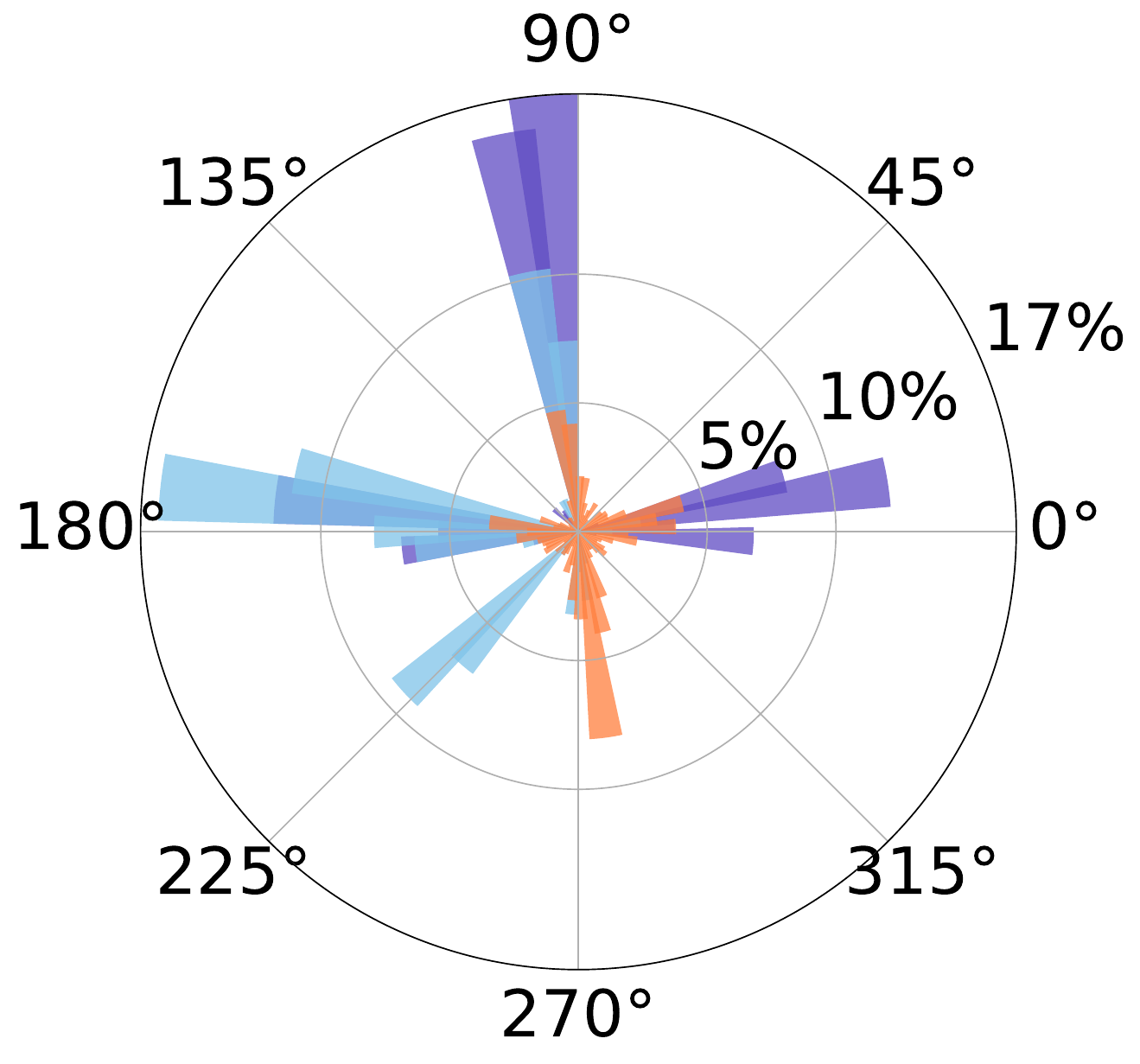}
    \caption{Yaw Angle}
    \end{subfigure}%

    \vspace{-8px}
    \caption{\small \textbf{Distribution of bounding box dimensions and yaw angles.} Vehicles exhibit a wide range of sizes.}
    \label{fig:classes_size_distribution}

    \vspace{-12px}
\end{figure*}

\begin{figure}
    \centering
    \includegraphics[width=\linewidth]{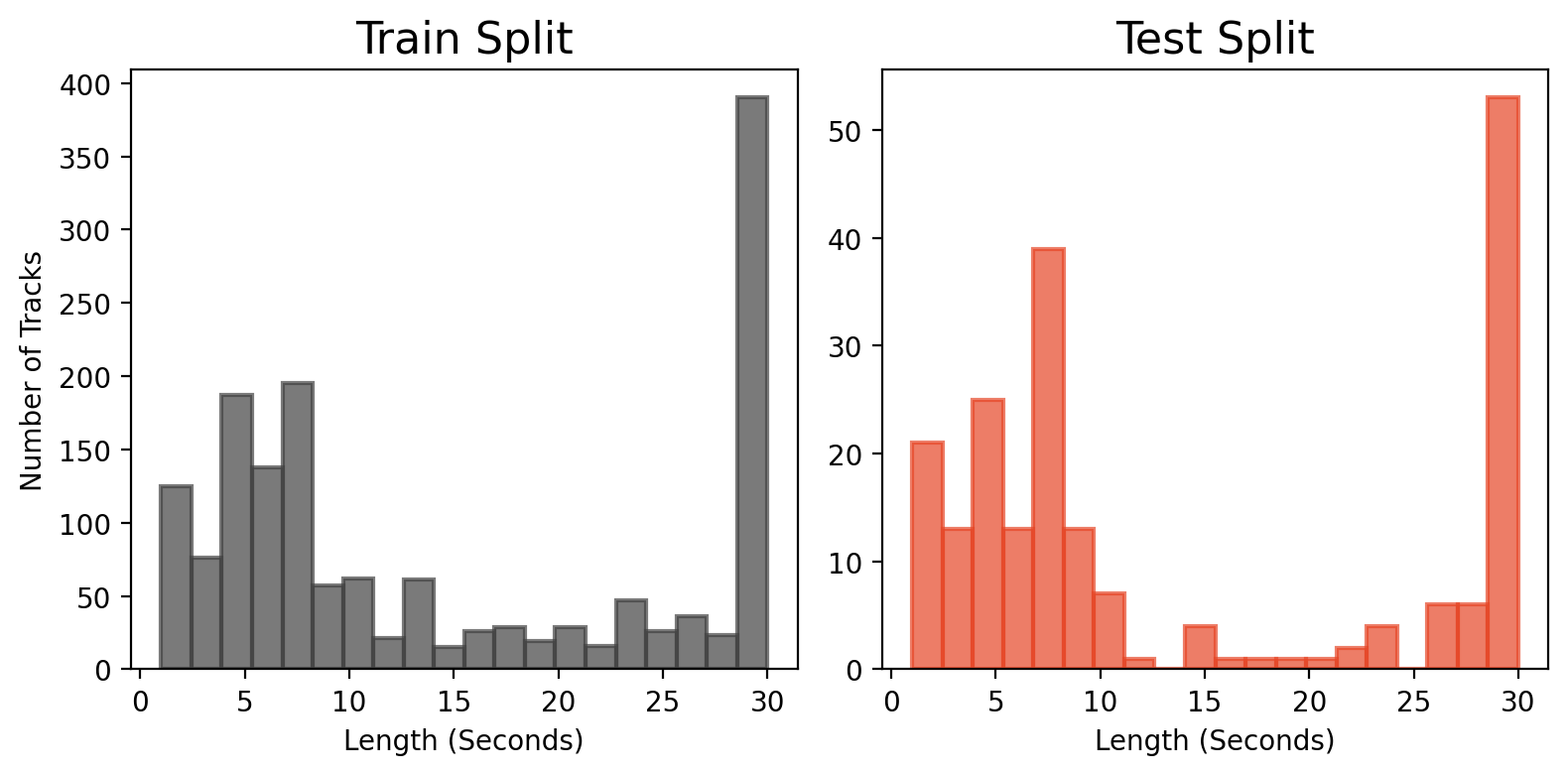}

    \vspace{-8px}
    \caption{\small \textbf{Distribution of track lengths.} The peak at 30 seconds corresponds to static objects.}
    \label{fig:track-length}

    \vspace{-12px}
\end{figure}

\begin{figure}
    \centering
    
    \begin{subfigure}{\linewidth}
        \centering
        \includegraphics[trim={1.4cm 0 1.4cm 0},clip,width=0.95\linewidth]{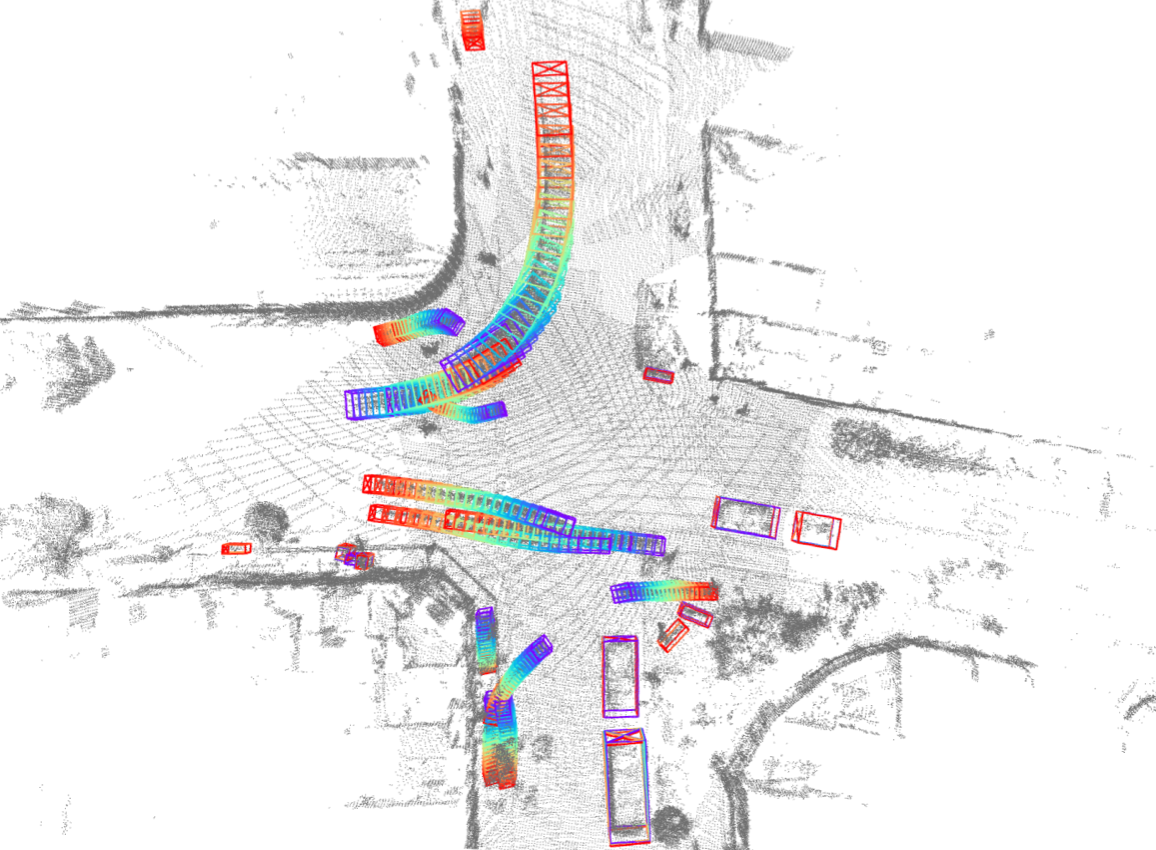}
    \end{subfigure}

    \begin{subfigure}{\linewidth}
        \centering
        \includegraphics[trim={0cm 0 0cm 0},clip,width=0.95\linewidth]{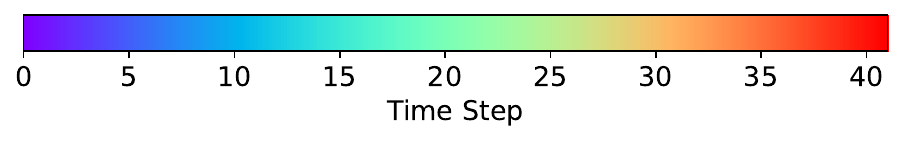}
    \end{subfigure}
    \caption{\small \textbf{Visualization of object tracks in \ours.} Dynamic objects display smooth trajectories, while static objects maintain consistent poses over time, highlighting the high quality of our annotations.}
    \label{fig:tracks-quality}

    \vspace{-10px}
\end{figure}

\mypara{Statistics.}
In our benchmark, we group 10 categories into 3 detection classes according to \autoref{tab:anno_cls_to_det_cls}. \autoref{fig:percentage_each_class_in_polar_coord} shows the distribution of annotations of three classes with respect to their polar coordinate in the coordinate system of \rsutop. \autoref{fig:classes_size_distribution} shows the distribution of dimensions and yaw angle of annotations of three classes. 
\autoref{fig:num_obj_per_class_in_train_test} shows the number of annotations of each class in the training set and test set.
\autoref{fig:track-length} analyzes track lengths in the training and test set. For both splits, most tracks are under 10 seconds. This is due to the dynamic and typical speeds at the intersection environment. A sharp peak at 30 seconds indicates the presence of static objects detected primarily by the RSU for the entire sequence duration.
\autoref{fig:tracks-quality} depicts the aggregation of point clouds from 5 agents and ground truth annotations in the coordinate system of \rsutop during a 4-second time span, which amounts to 40 time steps. 
The consistent pose of static objects and the smooth trajectory of dynamic objects visually demonstrate the quality of our annotation.

\begin{table}
\centering
\begin{tabular}{ @{} l  l @{}} 
\toprule
\textbf{Detection Class} & \textbf{Annotation Classes}  \\ 
\midrule
\multirow{2}{*}{Vehicle} & car, truck, emergency vehicle,   \\ 
                         & bus, electric vehicle, trailer \\
\midrule
\multirow{2}{*}{Bike} & motorbike, bicycle, \\
                      & portable personal mobility \\
\midrule
Pedestrian & pedestrian \\
\bottomrule
\end{tabular}

\vspace{-4px}
\caption{\small \textbf{Definition of detection classes.} The \ours dataset includes 10 fine-grained annotation classes for traffic participants, organized into 3 broader detection classes.} 

\label{tab:anno_cls_to_det_cls}
\end{table}

\section{Proposed Tasks and Benchmarks}
\begin{table*}[t]
\centering
\begin{tabular}{@{}lcclcclcclc@{}}
\toprule
 &  \multicolumn{2}{c}{Vehicle AP@} &  & \multicolumn{2}{c}{Bike AP@} &  & \multicolumn{2}{c}{Pedestrian AP@} &  & \multirow{2}{*}{\begin{tabular}[c]{@{}c@{}}Avg. Bandwidth\\ (MB)\end{tabular}} \\ \cmidrule(lr){2-3} \cmidrule(lr){5-6} \cmidrule(lr){8-9}
 &  IOU 0.5 & IOU 0.7 &  & IOU 0.5 & IOU 0.7 &  & IOU 0.3 & IOU 0.5 &  &  \\ \midrule
No Fusion &  0.42 & 0.42 &  & 0.19 & 0.19 &  & 0.47 & 0.41 &  & 0.00 \\
\midrule
\textcolor{Gray}{Early Fusion (Adapting \cite{xu2023v2v4real})} & \textcolor{Gray}{0.24} & \textcolor{Gray}{0.24} &  & \textcolor{Gray}{--} & \textcolor{Gray}{--} &  & \textcolor{Gray}{--} & \textcolor{Gray}{--} &  & \textcolor{Gray}{7.79} \\
Early Fusion & 0.65 & 0.65 &  & 0.65 & 0.65 &  & 0.74 & 0.67 &  & 7.79 \\
\midrule
Attentive Fusion~\cite{xu2022opv2v} & 0.82 & 0.82 &  & 0.71 & 0.71 &  & 0.74 & 0.68 &  & 5.26 \\
V2V-Net~\cite{wang2020v2vnet} & 0.72 & 0.72 &  & 0.69 & 0.69 &  & 0.42 & 0.32 &  & 4.19 \\
F-Cooper~\cite{chen2019f} & 0.75	& 0.75& & 0.68	& 0.68 & & 	0.72 &	0.65 & & 15.31\\
V2X-ViT~\cite{xu2022v2x} & 0.84 &	0.84 & & 0.71 & 	0.70 & & 0.77	& 0.70 & & 19.36 \\
V2V-AM~\cite{li2023learning} & 0.83 & 	0.83 & & 	0.79 & 	0.79	& & 0.69	& 0.60 & & 16.78 \\
where2comm~\cite{hu2022where2comm}  & 0.77 & 	0.77 & & 0.74 & 	0.74 & & 0.31	& 0.18 & & 16.78 \\ 
\midrule
Laly Fusion~\cite{dao2024practical} & 0.61 & 0.61 &  & 0.68 & 0.68 &  & 0.69 & 0.62 &  & 0.11 \\
\midrule
\textcolor{Gray}{Late Fusion (Adapting \cite{xu2023v2v4real})} & \textcolor{Gray}{0.12} & \textcolor{Gray}{0.12} &  & \textcolor{Gray}{--} & \textcolor{Gray}{--} &  & \textcolor{Gray}{--} & \textcolor{Gray}{--} &  & \textcolor{Gray}{0.11} \\
Late Fusion & 0.43 & 0.43 &  & 0.56 & 0.56 &  & 0.57 & 0.48 &  & 0.11 \\ \bottomrule
\end{tabular}
\caption{\small\textbf{Benchmarking results for the \textit{Collaborative Object Detection} task.}  All fusion methods outperform the No Fusion baseline, highlighting the advantage of collaborative perception. Each fusion method involves trade-offs between detection performance and communication bandwidth overhead. Models adapted from a premier R.H.S. V2V dataset \cite{xu2023v2v4real} are shown in \textcolor{gray}{gray}.}
\label{tab:v2x-default}
\end{table*}
\begin{table*}[ht]
\centering
\begin{tabular}{@{}lllcclcclcc@{}}
\toprule
                                                &                      &  & \multicolumn{2}{c}{Vehicle AP@}                 &  & \multicolumn{2}{c}{Bike AP@}             &  & \multicolumn{2}{c}{Pedestrian AP@}          \\ \cmidrule(lr){4-5} \cmidrule(lr){7-8} \cmidrule(l){10-11} 
                                                &                      &  & IOU 0.5              & IOU 0.7              &  & IOU 0.5              & IOU 0.7              &  & IOU 0.3              & IOU 0.5              \\ \midrule
\multicolumn{1}{l|}{\multirow{5}{*}{\evonetable+ RSU}}   & No Fusion (\evonetable only) & & 0.33	& 0.33 & & 0.28	& 0.28 & & 0.37	& 0.30 \\
\multicolumn{1}{l|}{}  & No Fusion (RSU only) &  & 0.22         & 0.22         &  & 0.20           & 0.19           &  & 0.26             & 0.22         \\
\multicolumn{1}{l|}{}                           & Attentive Fusion  &  & 0.53                 & 0.53                 &  & 0.60                 & 0.59                 &  & 0.57                 & 0.45                 \\
\multicolumn{1}{l|}{}                           & V2V-Net              &  & 0.46                 & 0.46                 &  & 0.47                 & 0.47                 &  & 0.32                 & 0.21                 \\
\multicolumn{1}{l|}{}                           & Late Fusion          &  & 0.29                 & 0.29                 &  & 0.43                 & 0.43                 &  & 0.52                 & 0.41                 \\ \midrule
\multicolumn{1}{l|}{\multirow{5}{*}{\evtwotable+ RSU}}   & No Fusion (\evtwotable only) &  & 0.33         & 0.33         &  & 0.16           & 0.16           &  & 0.08             & 0.05     \\
\multicolumn{1}{l|}{}   & No Fusion (RSU only) &  & 0.24         & 0.24         &  & 0.20           & 0.19           &  & 0.26             & 0.23     \\
\multicolumn{1}{l|}{}                           & Attentive Fusion  &  & 0.56                 & 0.56                 &  & 0.56                 & 0.56                 &  & 0.40                 & 0.27                 \\
\multicolumn{1}{l|}{}                           & V2V-Net              &  & 0.52                 & 0.52                 &  & 0.49                 & 0.48                 &  & 0.27                 & 0.18                 \\
\multicolumn{1}{l|}{}                           & Late Fusion          &  & 0.41                 & 0.41                 &  & 0.49                 & 0.49                 &  & 0.43                 & 0.31                 \\ \midrule
\multicolumn{1}{l|}{\multirow{5}{*}{\lasertable+ RSU}}  &  No Fusion (\lasertable only) &  &   0.30     & 0.30  &  &   0.32        & 0.32 & & 0.46 & 0.44 \\
\multicolumn{1}{l|}{} & No Fusion (RSU only) &  &0.17         & 0.17         &  & 0.18           & 0.18           &  & 0.25             & 0.22 \\
\multicolumn{1}{l|}{}                           & Attentive Fusion  &  & 0.71                 & 0.71                 &  & 0.66                 & 0.65                 &  & 0.58                 & 0.50                 \\
\multicolumn{1}{l|}{}                           & V2V-Net              &  & 0.63                 & 0.63                 &  & 0.55                 & 0.54                 &  & 0.37                 & 0.27                 \\
\multicolumn{1}{l|}{}                           & Late Fusion          &  & 0.46                 & 0.46                 &  & 0.52                 & 0.51                 &  & 0.66                 & 0.57                 \\ \bottomrule
\end{tabular}
\caption{\small\textbf{Benchmarking results for the \textit{Object Detection Enhanced by Communication to RSU} task.} Communication between the agent and RSU generally improves performance compared to single-agent perception. Performance varies across agents with different sensor configurations, suggesting future research opportunities to develop methods that work effectively with diverse sensor types.}
\label{tab:v2x-freeze-rsu}

\vspace{-6px}
\end{table*}

Our dataset includes multiple agents and annotations in the form of 3D bounding boxes with track IDs. This enables the development of methods for various collaborative perception tasks, such as object detection, tracking, and motion forecasting. Given the importance of object detection in autonomous driving, 
we focus on collaborative detection tasks in the main text and report preliminary tracking benchmark results in Appendix~\ref{sec:apdx-tracking-bench}.

\subsection{Definition of Tasks}
We define two tasks that are distinguished by the collaboration setting: \textit{Collaborative Object Detection} and \textit{Single-Vehicle Object Detection enhanced by communication to RSU}, which we describe in the following sections.

\mypara{Collaborative Object Detection.} 
This is the classical collaborative object detection task \cite{wang2020v2vnet, li2021learning}, where every connected agent (i.e., vehicles and RSUs) uses a shared model to (i) extract features from their point clouds, (ii) generate messages to send to other agents, and (iii) fuse the features of their point clouds with messages received from others.
The goal is to detect every visible object in a region of interest. 
We define visibility by comparing the number of \lidar points contained within an object's bounding box to a threshold. In this task, these \lidar points are sourced from any agents present within the region of interest.

\mypara{Object Detection Enhanced by Communication to RSU.}
This task assumes that the RSU model is designed and trained by a different provider than the one responsible for the CAVs' models.
In this task, the RSU model is pre-trained in the single-vehicle detection setting to detect objects visible to its LiDARs. After the pre-training process, the RSU model is fixed. CAVs in the proximity of the RSU receive messages from the RSU to enhance their detection capabilities. The objective is to detect all objects in a region of interest that are visible to either the CAV or the RSU.

The differences between this task and \textit{Collaborative Object Detection} are twofold. 
First, there is no communication among connected vehicles in this task, making it similar to Vehicle-to-Infrastructure (V2I) detection \cite{yu2022dair, zimmer2024tumtraf,xiang2024v2x}. 
Second, instead of having a single model shared among all connected agents like prior works on V2I collaboration, we have one model for the CAVs and another independent model for the RSU. 
This introduces a different challenge, as the CAV's model must adapt to messages from the RSU, which may contain domain gaps due to differences in model architecture, types of \lidar, and viewpoints.

\subsection{Benchmark}
\mypara{Evaluation Settings.} Since the annotations are made in the coordinate system of \rsutop, we define the region of interest for the two detection tasks as the range $\left[-51.2, 51.2\right]$ meters along both the x and y axes of this coordinate system.
For evaluation, we transform objects detected by each agent into this coordinate system. The visibility threshold is set to 5 points.
Since timestamp mismatches and localization errors are inherent in real-world applications and consequently present in our dataset, we do not artificially introduce them into the messages exchanged among connected agents (something that is often done in synthetic datasets \cite{xu2022opv2v, xu2022v2x}).
We measure object detection performance using Average Precision (AP). Detected objects are matched with ground truth based on their Intersection over Union (IoU) in the bird's-eye view plane. A detection and a ground truth object are considered a match if their IoU exceeds thresholds of 0.3, 0.5, or 0.7.
In addition to AP, we measure the bandwidth consumption of each collaborative method to gauge their practicality. The total bandwidth consumption is calculated by multiplying the number of agents in the collaboration network by the size of the message each agent sends. 
While the number of agents is not dependent on the collaboration method of choice, the message size is.
We report the bandwidth consumption by averaging the size of the messages that agents send, measured in Megabytes (MB). 
While some intermediate collaboration methods \cite{wang2020v2vnet, li2021learning, xu2022v2x} employ specialized compressing algorithms to reduce the message size, other methods \cite{xu2022opv2v, lu2023robust, dao2024practical} do not.
For fair comparison, we report uncompressed sizes.

\mypara{Methods.}
Our benchmark covers three conventional collaboration frameworks, namely Early fusion, Intermediate fusion \cite{wang2020v2vnet, xu2022opv2v, chen2019f, xu2022v2x, li2023learning, hu2022where2comm}, and Late fusion, and the recent \textit{Laly} fusion \cite{dao2024practical}. We detail the benchmarking methodology specifics in the appendix.

\subsection{Results}

\subsubsection{Collaborative Object Detection}
We show the benchmark of the \textit{Collaborative Object Detection} task in \autoref{tab:v2x-default}.
The results in this table clearly demonstrate the advantage of collaboration perception over single-agent perception, as all fusion methods largely outperform No Fusion on every class.
The comparison of three conventional fusion methods, including Early, Intermediate, and Late, shows that a higher precision is attained at the cost of a larger bandwidth consumption.
In contrast, \textit{Laly} fusion achieves comparable precision on Bike and Pedestrian compared to Early Fusion and Intermediate Fusion while consuming an order magnitude less bandwidth.
The high performance at less bandwidth of \textit{Laly} fusion coupled with its simplicity make this method a strong candidate for real-world deployment. 
However, we note that there is still ample room for improvement, particularly among the VRUs, suggesting the need for future algorithm design. 

\begin{figure}
    \centering
    \includegraphics[width=\columnwidth]{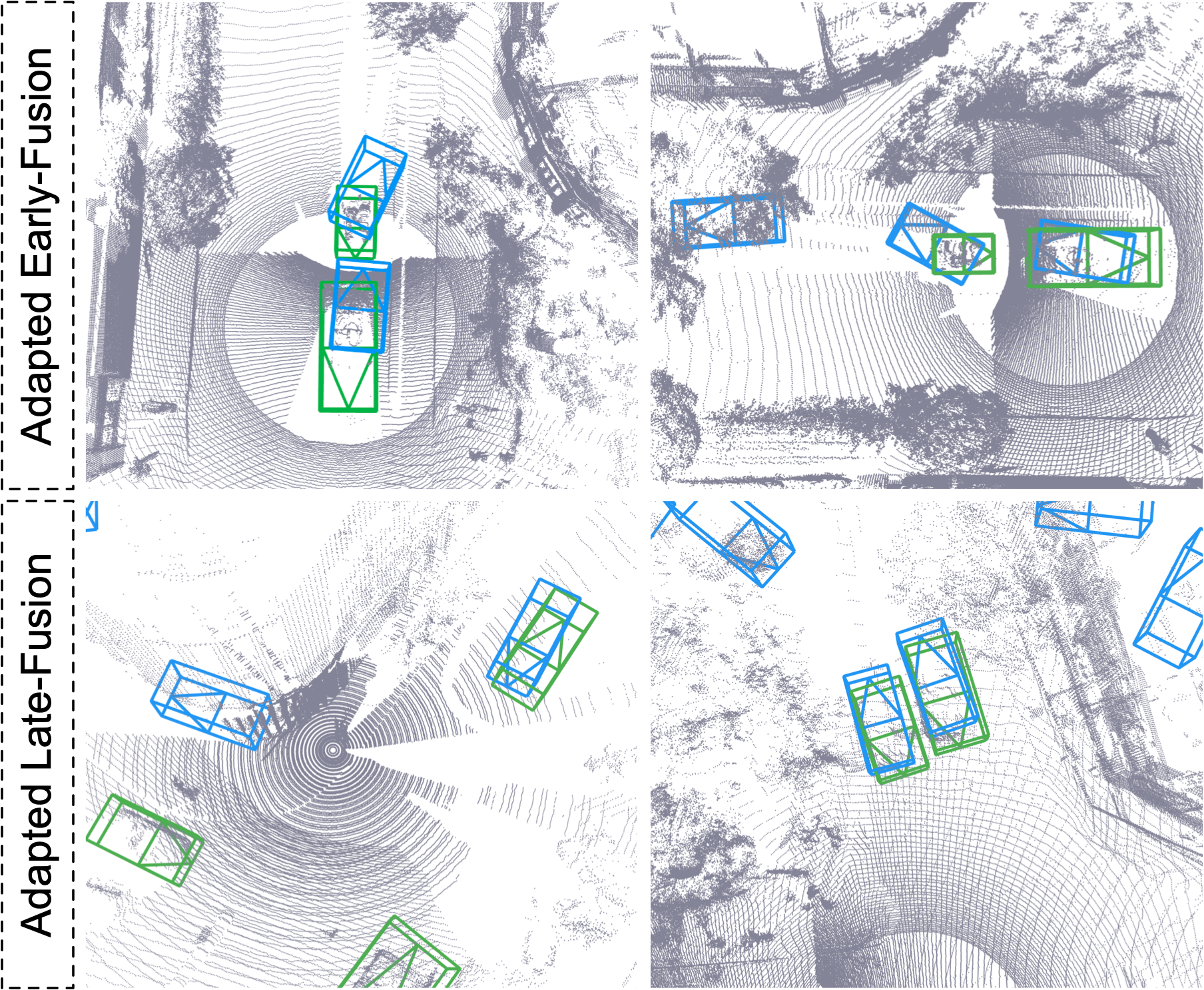}

    \caption{\textbf{Detection visualization} from adapting models trained on V2V4Real \cite{xu2023v2v4real} into \ours{}. Ground truth bounding boxes are shown in \textcolor{ForestGreen}{green} and predicted detections in \textcolor{CornflowerBlue}{blue}, with heading indicated by the triangle. Observe that predicted heading directions are often aligned to priors learned in R.H.T. driving.}

    \label{fig:adapt-detect}

    \vspace{-12px}
\end{figure}

\mypara{Domain Gap from prior R.H.T. Datasets.}
To illustrate the domain gap covered by \ours{}, we directly adapt an early-fusion and a late-fusion model trained on the right-hand traffic (R.H.T.) dataset, V2V4Real \cite{xu2023v2v4real}, onto our dataset (grayed-out rows in \autoref{tab:v2x-default}). 
Performance degraded significantly, with vehicle headings predicted incorrectly, indicating a learned prior from traffic flow (\autoref{fig:adapt-detect}). This highlights a substantial domain gap due to left-hand traffic and sensor modality differences, underscoring Mixed Signals' unique contribution to the V2X perception landscape.

\subsubsection{Detection via Communication to RSU}
\autoref{tab:v2x-freeze-rsu} presents the performance of different fusion methods on \textit{Object Detection Enhanced by Communication to RSU} task. 
In this setting, detector training is more challenging, as each vehicle-centric detector must adapt to a frozen RSU detector. Nevertheless, results show that communication with RSU is still advantageous, as evidenced by the substantial performance improvement over the No Fusion baselines.
Furthermore, the performance of the Laser car is better than the performance of the two EVs.
This is because the \lidar of the Laser car has a 360-degree coverage of its surroundings.
On the other hand, the tilted angle of the \lidar on the two EVs makes the region behind them unobservable.
The {\lidar}s on the two EVs do not capture the intensity information, resulting in a domain gap between their features and those from the RSU.
These observations point to future research directions for developing methods that could work well with diverse sensor configurations.

\section{Discussion and Conclusion}

Our work presents the \ours{} V2X dataset, created through careful data selection, sensor synchronization and localization, and a strong investment in high quality annotations. 
To the best of our knowledge, our dataset is the first to support heterogeneous sensor configurations with varying positions and orientations, collected in an out-of-domain left-hand traffic country, Australia, providing a diverse dataset addition to the field.
We hope that the release of our dataset will facilitate research into 
complex and realistic settings for V2X perception.
Future directions of research include studying communication protocols that ensure both fast transmission 
and directed communication that targets salient information.

\section*{Acknowledgement}
We thank Runsheng Xu and Hao Xiang for their insightful discussions and support throughout the early stage of this project. This research is funded by University of Sydney – Cornell University Ignition Grants/Global Strategic Collaboration Awards, National Science Foundation (IIS-2107161), and the New York Presbyterian Hospital.
Minh-Quan Dao is funded by ANNAPOLIS project managed by the French National Agency for Research (ANR-21-CE22-0014), and Katie Luo by AAUW American Dissertation Fellowship.
 
{
    \small
    \bibliographystyle{ieeenat_fullname}
    \bibliography{main}
}

\clearpage
\setcounter{page}{1}
\maketitlesupplementary

\renewcommand{\thesection}{\Alph{section}}
\renewcommand{\thesubsection}{\thesection.\arabic{subsection}}
\renewcommand{\thetable}{A\arabic{table}}
\renewcommand{\thefigure}{A\arabic{figure}}
\renewcommand{\theequation}{A\arabic{equation}}
\setcounter{figure}{0}
\setcounter{table}{0}
\setcounter{equation}{0}
\setcounter{section}{0}

\noindent
In this appendix material, we include: 1) extra details about the Mixed Signals dataset and the provided code devkit,
2) annotation details and instructions given to annotators, and 3) additional sensor details. We include an additional dataset teaser video in the dataset website\footnote{\url{https://sites.coecis.cornell.edu/mixedsignals/\#introvid}} that we encourage readers to watch.

\begin{table*}[!ht]
\centering

\resizebox{\linewidth}{!}{%
\begin{tabular}{@{}ll@{}}

\toprule
{ \textbf{Category}}                                      & \textbf{Definition}                                                                                                                                                                               \\ \midrule
Car                                                                           & \begin{tabular}[c]{@{}l@{}}Includes passenger vehicles such as sedans, hatchbacks, SUVs, and coupes that are designed primar-\\ ily for the transportation of  passengers.\end{tabular}                                               \\ \midrule
Truck                                                                         & \begin{tabular}[c]{@{}l@{}}Encompasses larger vehicles primarily used for transporting goods and materials. This category  \\includes pickup trucks, delivery trucks, and heavy-duty trucks.\end{tabular}                           \\ \midrule
Emergency Vehicle                                                             & \begin{tabular}[c]{@{}l@{}}Vehicles designated for emergency response, including ambulances, fire trucks, police cars, and other \\ vehicles equipped with sirens and emergency lights.\end{tabular}                                \\ \midrule
Bus                                                                           & \begin{tabular}[c]{@{}l@{}}Large motor vehicles designed to carry numerous passengers. Buses include city transit buses, school \\buses, and intercity coaches. They usually have designated routes and schedules.\end{tabular} \\ \midrule
\begin{tabular}[c]{@{}l@{}}Motorcycle\\ Motorized Bike\end{tabular}           & \begin{tabular}[c]{@{}l@{}}Two-wheeled motor vehicles, including motorcycles and motorized bikes. This category also \\covers scooters and mopeds.\end{tabular}                                                                       \\ \midrule
\begin{tabular}[c]{@{}l@{}}Portable Personal \\ Mobility Vehicle\end{tabular} & \begin{tabular}[c]{@{}l@{}}Small, lightweight vehicles designed for personal mobility, including electric scooters, hoverboards,\\  and segways.\end{tabular}                                                                           \\ \midrule
Bicycle                                                                       & \begin{tabular}[c]{@{}l@{}}Human-powered, pedal-driven vehicles with two wheels. Bicycles include standard bikes,  mountain \\ bikes, and road bikes. This category include motorized bicycles or electric bikes.\end{tabular}       \\ \midrule
Electric Vehicle                                                              & Refers to small, golf car-like vehicles used for data collection purposes.                                                                                                                                                             \\ \midrule
Trailer                                                                       & Non-motorized vehicles designed to be towed by a motor vehicle.                                                                                                                                                                        \\ \midrule
Pedestrian                                                                    & \begin{tabular}[c]{@{}l@{}}Individuals traveling on foot. This category includes people walking or running.\end{tabular}                                                                                                            \\ \bottomrule

\end{tabular}
}
\caption{\textbf{Definitions of the annotation classes.}}
\label{table:v2x_label}
\end{table*}

\section{Data and Devkit}
Please see \url{https://mixedsignalsdataset.cs.cornell.edu/} for the dataset download instructions and the provided devkit. Below, we add a brief description of the devkit and visualize a dataset sample.

\subsection{Devkit Description}
We provide a separate devkit and additionally integrate our dataset into the framework OpenCOOD \cite{xu2022opv2v}, which offers the implementation of various state-of-the-art collaborative perception methods. 
As OpenCOOD only provides single-class models, we adapt its implementation of Early, Intermediate, and Late Fusion models to detect three classes, including vehicles, bikes, and pedestrians.
We added detection heads of 1-by-1 convolution layers to existing architectures to achieve this.
In addition, we add the recent \textit{Laly} fusion \cite{dao2024practical} to this framework.
Every model in our benchmark uses PointPillar \cite{lang2019pointpillars} as the backbone.
Interested readers can refer to our devkit\footnote{\url{https://github.com/quan-dao/mixed-signals-devkit}} and code release\footnote{\url{https://github.com/acfr/Mixed-Signals-Dataset}} and extended OpenCOOD integration for further details on architectures and training settings. 

\subsection{Sample Data}
\begin{figure}[h!]
\centering
\includegraphics[width=\columnwidth]{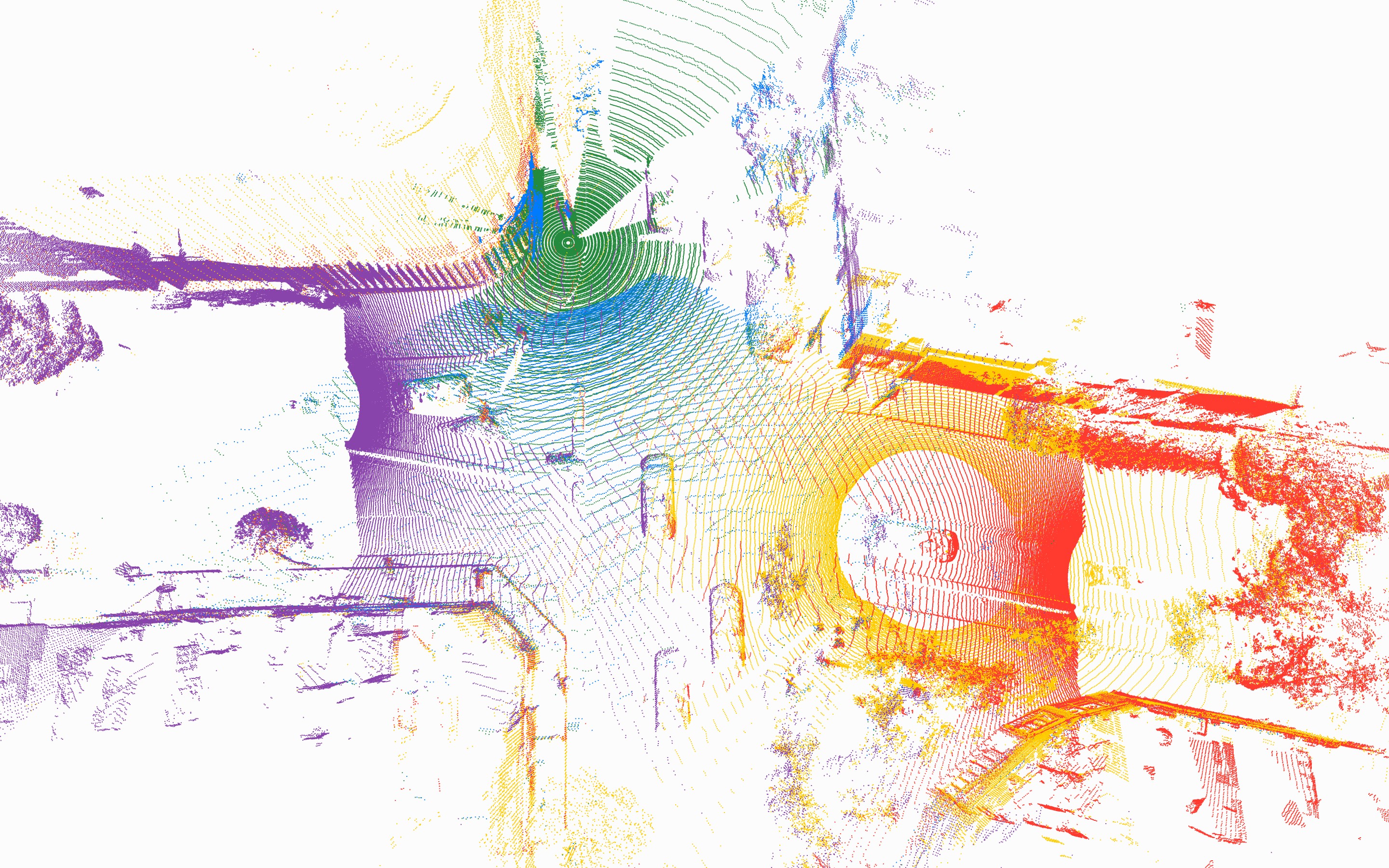}
\caption{\textbf{Top-down view of the data collected at the location.} \lidar point clouds are colored by the vehicle and RSU that collected them, consisting of the 3 vehicle agents ({\color{red}red}, {\color{Goldenrod}yellow}, and {\color{Purple}purple}) and the Top and Dome LiDAR sensors of the RSU ({\color{ForestGreen}green}, {\color{blue}blue}). Best viewed in colour.}
\label{fig:data_collection}
\end{figure}

\autoref{fig:data_collection} shows an example of the collected data, where the points are colour-coded to represent the different {\lidar}s. 
The dataset aims to replicate realistic urban scenarios that reflect the complexities of real-world implementations by using multiple vehicles with diverse sensor configurations and a roadside unit. 
Real-world deployments of autonomous vehicles on streets incorporate {\lidar}s, which are becoming more affordable. Roadside infrastructure, such as roadside units, is also gaining popularity for traffic monitoring and data analytics, now often equipped with LiDAR, traffic light timing information, and communication systems to enhance robustness and applicability.
Our dataset consists of \lidar point clouds, which offer the advantage of not capturing identifiable information like faces or license plates, thus preserving data privacy. This contrasts with camera images, which often require post-processing to anonymize sensitive details, potentially affecting data quality. Our dataset includes tracking IDs for each bounding box, and this information will be released alongside this paper. Benchmarks will be made available at a later date.

\begin{figure}[t!]
\centering
\includegraphics[width=\columnwidth]{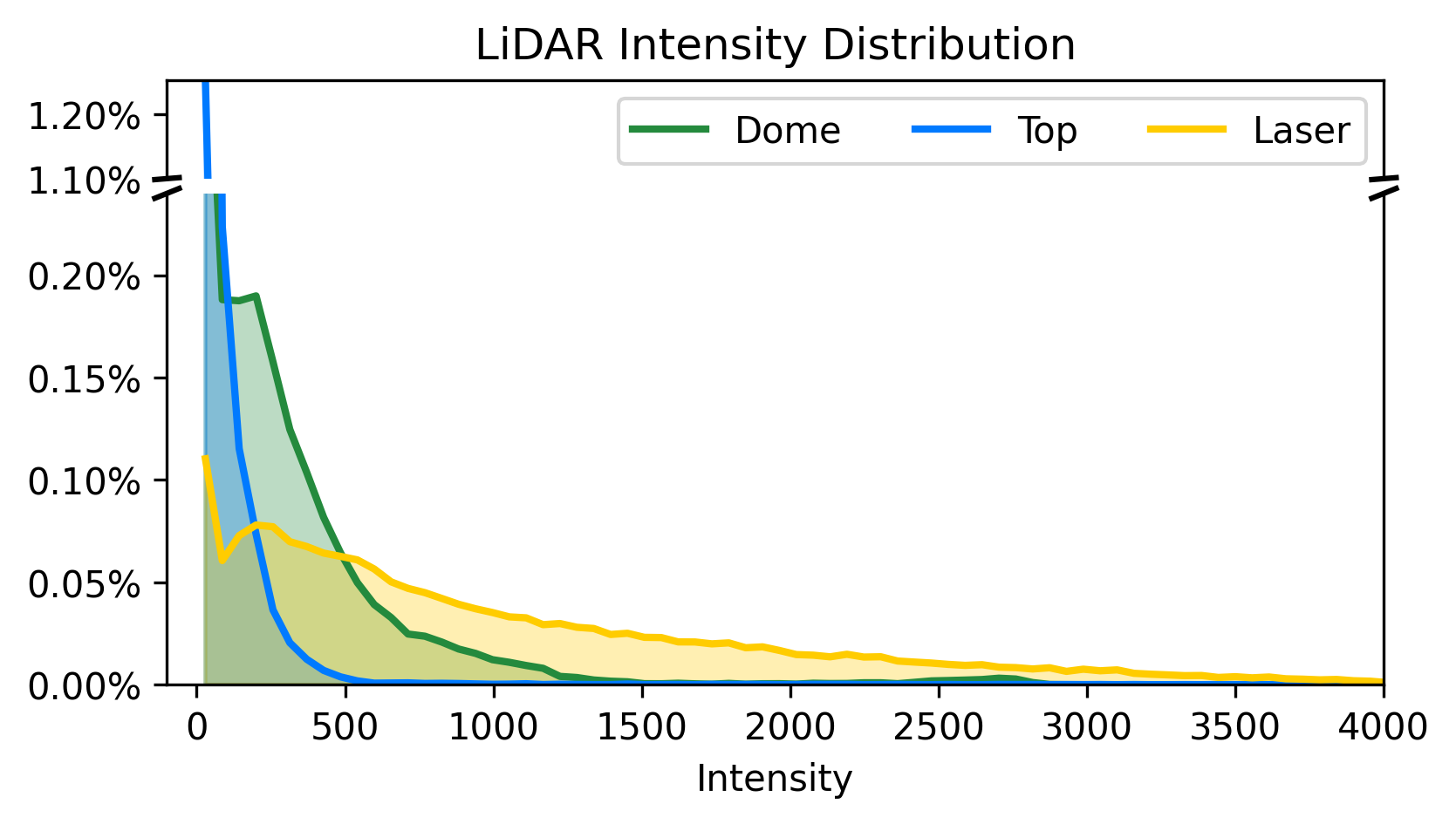}

\vspace{-10px}
\caption{\small \textbf{Distribution of \lidar intensities from RSU \rsutop, \rsudome, and \laser car sensors.} Each sensor shows different intensity ranges and distributions. \evone and \evtwo \lidar sensors do not have intensity readings.}
\label{fig:lidar-intensities}

\vspace{-16px}
\end{figure}

\mypara{Intensity Distributions.}
\autoref{fig:lidar-intensities} shows LiDAR intensity distributions from RSU \rsutop, \rsudome, and \laser car sensors. 
\rsudome and \rsutop sensors record higher intensities because there is a large number of static objects (e.g., buildings, traffic lights) near them.
In contrast, the \laser car sensor presents a smoother decline in intensity values because of its location on the vehicle, which allows the detection of objects at greater distances. 
EV-1 and EV-2 sensors do not capture intensity readings. 
Therefore, a uniform approach to utilizing intensity values across all agent models would be inadequate.

\section{Annotation Instructions}
\label{sec:apdx-annotation-instructions}
We provide the instructions given to the Segments.ai\footnote{\url{https://segments.ai/}} annotators in the attached material, titled ``\textit{Spec Document - Multi-sensor labeling}" at the bottom of the appendix.
We selected to invest in the quality of the annotations,
applying rigorous quality control measures to guarantee accurate and consistent labeled data, minimizing errors, and maintaining high standards.

\subsection{Definitions of the Annotation Classes}
The Mixed Signals dataset categories road agents in different vehicle types and pedestrians. Categories such as ``Car" and ``Truck" encompass common passenger and large transport vehicles, while ``Emergency Vehicle" covers ambulances, fire trucks, and police cars, highlighting their importance in urban scenarios. ``Bus" labels are designated for large passenger vehicles typically used in public transportation. The dataset also distinguishes between ``Motorcycle" and ``Motorized Bike," and ``Portable Personal Mobility Vehicle," which includes modern personal transport devices like electric scooters and hoverboards. Traditional ``Bicycle" labels account for both standard and electric bikes. Labels for ``Electric Vehicle" and ``Trailer" ensure that smaller, often data-collection vehicles and towable units are accurately represented. Finally, we labeled humans as ``Pedestrians".
In \autoref{table:v2x_label}, we provide the definitions of the 10 fine-grained annotation classes in the \ours dataset. The breakdown of the fine-grain classes into the benchmarked classes can be found in the main text.

\subsection{Track Annotations and Multi-agent Tracking}
\label{sec:apdx-tracking-bench}
We benchmark the performance of the planned tracking task for our dataset. The Mixed Signals dataset has labels for track ID's, as seen in Figure 10 of the main text. We hope to include and benchmark tracking methods as an additional task which is supported by our dataset. We report some initial benchmarking results on the AB3DMOT tracking method \cite{Weng2020_AB3DMOT} in \autoref{tab:tracking}. For further details about track labels, please explore the data itself; a distribution of the tracks are in Figure 9 of the main text.

\begin{table}[t]
\centering
\begin{tabular}{@{}lccc@{}}
\toprule
Category & sAMOTA & AMOTA & AMOTP \\ \midrule
Vehicle & 89.6 & 43.1 & 63.3 \\
Pedestrian & 76.6 & 32.6 & 42.8 \\ \bottomrule
\end{tabular}

\vspace{-4px}
\caption{Tracking performance for AB3DMOT with V2X-ViT detections on the Mixed Signals validation split.}
\label{tab:tracking}

\vspace{-8px}
\end{table}

\subsection{Annotation Details}
The annotation process for this multi-sensor dataset involves handling joint scenes and synchronization discrepancies between sensors. Due to time synchronization, fast-moving objects might appear slightly offset across the data collected from different sensors. To address these discrepancies, annotators were instructed to prioritize the roadside unit point cloud for bounding box creation, following a set hierarchy. When there is a mismatch, bounding boxes should be aligned with the point cloud in the following order: roadside unit,  EVs, and the urban vehicle. For example, if there is a difference between the roadside unit and the vehicles' point cloud, the bounding box should only be fitted around the roadside unit points. This ensures consistency in object localization across frames despite synchronization lags.

\section{Sensor Details}

\subsection{Hardware and Synchronization Details}
\label{sec:apdx-hw-synchronization}

\begin{table}[th!]
\vspace{-8px}
\centering
\resizebox{\linewidth}{!}{
\begin{tabular}{l c c c c}
\toprule
\textbf{Sensor} & \textbf{Agent} & \textbf{Range$^*$} & \textbf{Channels} & \textbf{Vertical FOV} \\ \midrule
Ouster OS1-128         & Vehicles       & 170 m          & 128               & 45                    \\ 
Ouster OS1-64          & RSU            & 100 m          & 64                & 45                    \\ 
Ouster OS Dome         & RSU            & 45 m           & 128               & 180                   \\ \bottomrule
\multicolumn{5}{l}{\small $^*$Based on 80\% Lambertian reflectivity in the sensors' official datasheets.} \\
\end{tabular}}

\vspace{-4px}
\caption{\small\textbf{Hardware specifications.} }
\label{tab:sensor-details}

\vspace{-8px}
\end{table}

\noindent
Synchronization is especially important in dynamic environments, as any introduced time shifts can lead to positional inconsistencies, resulting in multiple detections of the same object.
The sensors in our multi-agent system were timestamped using GPS time as a common reference, and sensor details are provided in \autoref{tab:sensor-details}.
Rotational LiDARs continuously scan the environment in 360 degrees, thus, different portions of the surroundings are captured at slightly different times during a full rotation. When vehicles are in motion, their positions and orientations change dynamically between LiDARs sweeps. 
The maximum time gap for matching sensor readings between 10 Hz rotational sensors is 50 ms. 
Since sensors rotate fully in 100 ms, angular positions differ by at most 180 degrees.
If the time difference between readings were larger than 50 ms, it would be 
matched with the next or previous rotation instead.
As shown in the original manuscript, precise sensor synchronization, robust multi-agent localization, and clearly defined annotation protocols produced high-quality data association across all sensors.

\subsection{Localization}
\label{sec:apdx-localization}

Localization is one of the most critical tasks for CAV, estimating their position relative to a global reference frame. 
One of the most commonly used sensors for localization is the Global Navigation Satellite System (GNSS). GNSS offers access to a satellite constellation that provides global positioning via triangulation. However, despite its widespread use, GNSS has several drawbacks, particularly in urban environments. Its accuracy can be reduced in urban canyons, where tall buildings block or reflect signals, leading to degraded positioning accuracy. To overcome this problem, we use dense and accurate point cloud maps \cite{liosam2020shan} as references for our localization algorithm.

\subsection{Definition of Heterogeneity in Sensor Suite}
Heterogeneity in our context refers to the variability between LiDAR sensors and platform geometry within a single dataset. Heterogeneity can appear in multiple forms \cite{jung2023helipr}; our dataset represents it in five LiDARs that span three models, each mounted in four configurations. In line with the feedback, Tab. 1 of the original manuscript has been updated accordingly.
Our dataset demonstrates a realistic setting where collaborative agents have different LiDAR models and position them in different configurations. 

\includepdf[pages=-]{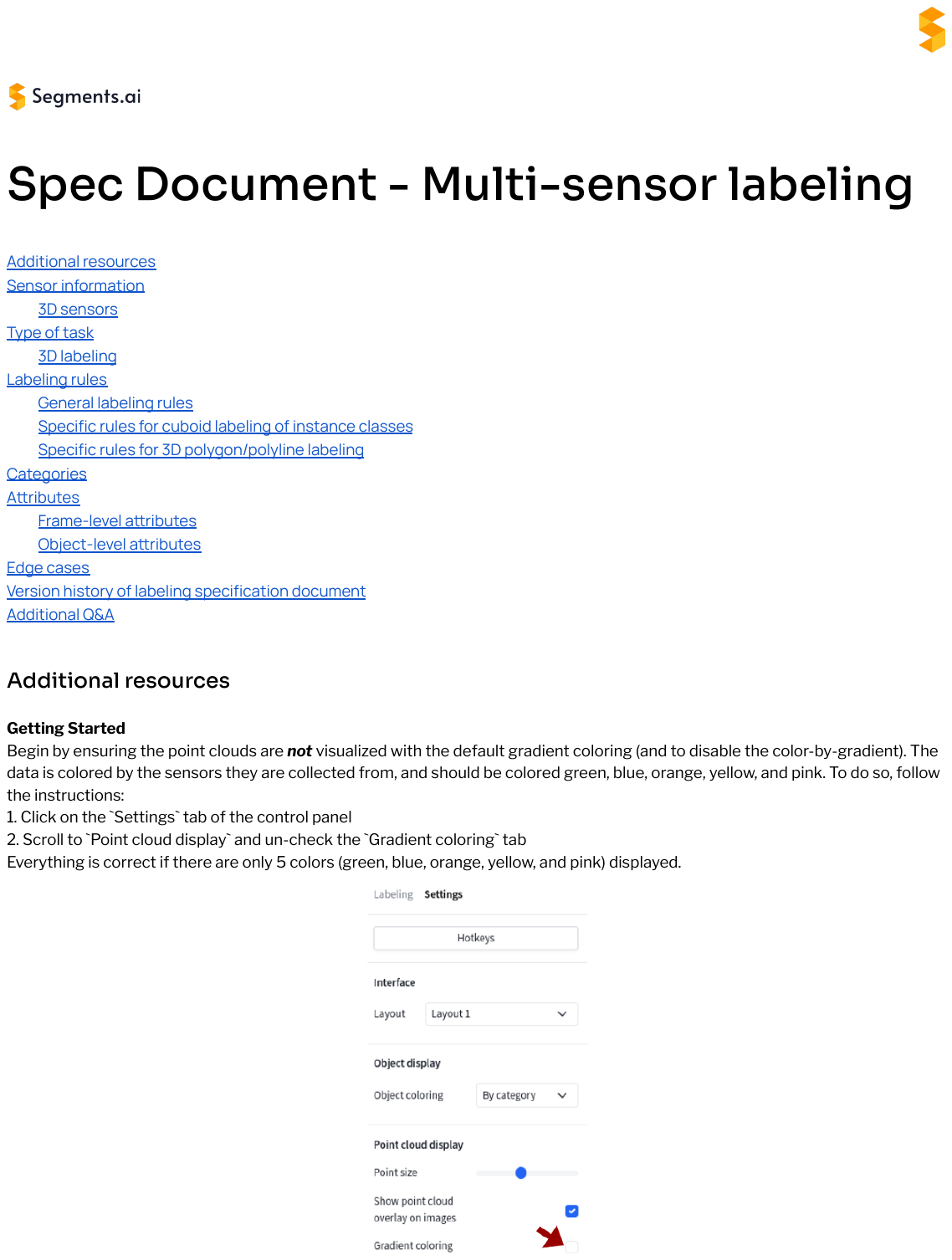}

\end{document}